# PA-LVIO: Real-Time LiDAR-Visual-Inertial Odometry and Mapping with Pose-Only Bundle Adjustment

Hailiang Tang, Tisheng Zhang, Liqiang Wang, Xin Ding, Man Yuan, and Xiaoji Niu

***Abstract*—Real-time LiDAR-visual-inertial odometry and mapping is crucial for navigation and planning tasks in intelligent transportation systems. This study presents a pose-only bundle adjustment (PA) LiDAR-visual-inertial odometry (LVIO), named PA-LVIO, to meet the urgent need for real-time navigation and mapping. The proposed PA framework for LiDAR and visual measurements is highly accurate and efficient, and it can derive reliable frame-to-frame constraints within multiple frames. A marginalization-free and frame-to-map (F2M) LiDAR measurement model is integrated into the state estimator to eliminate odometry drifts. Meanwhile, an IMU-centric online spatial-temporal calibration is employed to obtain a pixel-wise LiDAR-camera alignment. With accurate estimated odometry and extrinsics, a high-quality and RGB-rendered point-cloud map can be built. Comprehensive experiments are conducted on both public and private datasets collected by wheeled robot, unmanned aerial vehicle (UAV), and handheld devices with 28 sequences and more than 50 km trajectories. Sufficient results demonstrate that the proposed PA-LVIO yields superior or comparable performance to state-of-the-art LVIO methods, in terms of the odometry accuracy and mapping quality. Besides, PA-LVIO can run in real-time on both the desktop PC and the onboard ARM computer. The codes and datasets are open sourced on GitHub (https://github.com/i2Nav-WHU/PA-LVIO) to benefit the community.**

## Nomenclature

| *Variables and Symbols* | |
|---|---|
| $\mathbf{q}, \mathbf{R}, \boldsymbol{\phi}$ | Attitude quaternion, rotation matrix, and rotation vector |
| $\otimes$ | Quaternion product |
| Log, Exp | Transformation between the quaternion and rotation vector |
| $\boldsymbol{p}_{\mathrm{wb}}^{\mathrm{w}}, \mathbf{q}_{\mathrm{b}}^{\mathrm{w}}$ | The IMU pose (body frame, b-frame) w.r.t the world frame (w-frame) |
| c, u | Camera frame (c-frame) and normalized camera frame (u-frame) |
| r | LiDAR frame (r-frame) |
| $\boldsymbol{p}_{\mathrm{bc}}^{\mathrm{b}}, \mathbf{q}_{\mathrm{c}}^{\mathrm{b}}$ | Camera-IMU extrinsic parameters |
| $\boldsymbol{p}_{\mathrm{br}}^{\mathrm{b}}, \mathbf{q}_{\mathrm{r}}^{\mathrm{b}}$ | LiDAR-IMU extrinsic parameters |
| $\boldsymbol{p}_{\mathrm{rc}}^{\mathrm{r}}, \mathbf{q}_{\mathrm{c}}^{\mathrm{r}}$ | Camera-LiDAR extrinsic parameters |
| $\boldsymbol{v}$ | Velocity vector |
| $\delta t$ | Time-delay parameter |
| $\boldsymbol{e}$ | Measurement residual |
| $\boldsymbol{\Sigma}$ | Measurement covariance |

| *Abbreviations* | |
|---|---|
| BA | Bundle adjustment |
| F2F | Frame to frame |
| F2M | Frame to map |
| FGO | Factor graph optimization |
| LIO | LiDAR-inertial odometry |
| LVIO | LiDAR-visual-inertial odometry |
| MSC | Multi-state constraint |
| PA | Pose-only bundle adjustment |
| VIO | Visual-inertial odometry |

This work was supported in part by the National Natural Science Foundation of China under Grant 42504027 and 42374034, the China Postdoctoral Science Foundation under Grant Number 2025M780217, and the Key Research and Development Program of Hubei Province under Grant 2024BAB024. (*Corresponding author: Tisheng Zhang.*)

Hailiang Tang, Tisheng Zhang and Xiaoji Niu are with the GNSS Research Center, Wuhan University, Wuhan 430079, China, the Hubei Technology Innovation Center for Spatiotemporal Information and Positioning Navigation, Wuhan 430079, China, and the Hubei Luojia Laboratory, Wuhan 430079, China. (e-mail: thl@whu.edu.cn; zts@whu.edu.cn; xjniu@whu.edu.cn)

Liqiang Wang, Xin Ding, and Man Yuan are with the GNSS Research Center, Wuhan University, Wuhan 430079, China (e-mail: wlq@whu.edu.cn; dingxin@whu.edu.cn; yuanman@whu.edu.cn).

## I. Introduction

REAL-TIME mapping or reconstruction not only can be used for immediate visualization, but also can be employed for navigation and planning in intelligent transportation systems [1]. Hence, building a high-quality map has become an essential capability for autonomous robots, unmanned aerial vehicles (UAV), and portable mapping systems. Typically, RGB cameras are adopted for mapping tasks, due to the lower cost and satisfactory accuracy. Besides, the visual-inertial system [2] can achieve 3-dimensional (3-D) reconstruction using a single camera, by incorporating a low-cost micro-electro-mechanical system (MEMS) inertial measurement unit (IMU) [3]. However, dense reconstruction using visual sensors is difficult and complex, especially in large-scale scenarios. Recently, (half-)solid-state 3-D light detection and ranging (LiDARs) with reduced cost and long-distance measurements have been manufactured, and they can be employed for dense 3-D reconstruction in large-scale environments.

By integrating visual and LiDAR sensors [4], [5], [6], [7], a dense, RGB-colored, point-cloud map can be obtained, which can be further processed for tasks such as navigation and planning [1]. During the mapping process, the scene can be reconstructed using LiDAR point clouds and can be rendered using visual textures. Besides, the MEMS IMU, which can provide high-rate prior pose and remove motion distortions of point clouds, also plays a key role in high-quality mapping. According to the measurement characteristics, the camera is an excellent rotation sensor, and LiDAR is an accurate translation sensor. Meanwhile, the high-rate IMU can serve as a spatial-temporal synchronization center, which is an important and effective link for multi-sensor fusion. Hence, LiDAR-visual-inertial fusion has gradually become an effective solution for low-cost and real-time 3-D mapping.

### *A. Tightly-Coupled or Loosely-Coupled LVIOs*

As the foundation of large-scale mapping, accurate and reliable pose estimation is usually derived from a LiDAR-visual-inertial odometry (LVIO) [4], [7], [8], while place recognition or loop closure can be further adopted to eliminate drifts [9], [10]. An LVIO can be separated into two subsystems, *i.e.*, LiDAR-inertial odometry (LIO) [11], [12], [13] and visual-inertial odometry (VIO) [2], [9], [14].

A loosely-coupled LVIO obtains the final pose by integrating pose results from the VIO and LIO subsystems in a pose-graph optimizer. VIRAL-Fusion [15], RTAB-Map [16], LVI-SAM [8], and Super Odometry [17] are all loosely-coupled LVIOs. In particular, R2LIVE [18] and LVIO-Fusion [19] incorporate the pose derived from LIO into the VIO estimator, and they are a kind of half-tightly-coupled LVIO. Similarly, Zhang [20] and VIL-SLAM [21] use the VIO pose as the initial state for LiDAR scan matching, so as to improve the LiDAR mapping accuracy. Nevertheless, these loosely-coupled LVIOs may result in degraded robustness in LiDAR and visual-degenerated environments.

Tightly-coupled LVIOs have been proven more robust and accurate than loosely-coupled LVIOs. LIC-Fusion [22], [23] tightly integrate LiDAR, visual, and inertial measurements within the multi-state constraint (MSC) Kalman filter (MSCKF) [24]. In contrast, a factor graph optimization (FGO) is employed for tightly-coupled LVIO [25]. Recently, the iterated extended Kalman filter (IEKF) has exhibited superior real-time performance in FAST-LIO [12], [26], which only employ the newest LiDAR frame to update the system states. R3LIVE [4], [5] employs the point-to-plane LIO from FAST-LIO2 [12] and a photometric VIO to update IMU states. Similarly, FAST-LIVO [6], [7] integrates a sparse-direct visual alignment into FAST-LIO2 [12], and constructs an efficient LVIO. Benefiting from the adopted tightly-coupled frameworks, R3LIVE [4], [5] and FAST-LIVO [6], [7] exhibit satisfactory robustness and accuracy.

### *B. F2F or F2M Constraints*

As a kind of dead-reckoning (DR) system, frame-to-frame (F2F) [11], [27] visual and LiDAR measurements should be employed in LVIOs. F2F measurements construct a kind of relative pose constraints, and thus they satisfy the characteristics of a consistent state estimator [27]. LIC-Fusion [22], [23] integrates F2F visual and LiDAR feature measurements into an MSCKF-based state estimator. In particular, LIC-Fusion 2.0 [23] employs a sliding-window plane-feature tracking to derive the F2F LiDAR measurements. Similarly, the tracked LiDAR plane and line features are adopted to construct F2F 3D plane and line measurements in an FGO-based state estimator [25]. F2F-based LVIOs can fully utilize measurements from multiple frames to achieve consistent state estimation. Nevertheless, they may drift gradually even in small-scale scenarios.

In contrast, frame-to-map (F2M) measurements construct a kind of absolute pose constraint between the current frame and the built map. The F2M constraint has originally been adopted in LIOs like LIO-SAM [28] and FAST-LIO2 [12], and it can eliminate odometry drifts without using any loop-closure method. As for LVIO, both R3LIVE [4], [5] and FAST-LIVO [6], [7] adopt F2M visual and LiDAR updates within the IEKF framework to perform accurate state estimation, especially in small-scale scenarios. However, F2M-based method has exhibited inconsistency in state estimation [11], mainly because the pose estimation is achieved by registering the newest frame to the previously built map. This inconsistency has several notable drawbacks: (1) A wrongly built map may further ruin the state estimator, *e.g.*, divergence in height direction [13], [29]; (2) The accuracy of the inertial navigation system (INS) [30] may be reduced because of incorrectly estimated accelerometer biases, which means misaligned attitude to the gravity direction [13], [29]; (3) The LiDAR-IMU or visual-IMU extrinsic parameters cannot be estimated due to the F2M registration, as the pose estimation is bound to the map [11]. Due to the above reasons, F2M-based methods may degrade accuracy, especially in large-scale scenarios.

In short, F2F-based LVIOs are consistent in terms of state estimation, while they may diverge over time. On the contrary, F2M-based LVIOs can eliminate drifts by employing F2M update, though they have some state-estimation problems. If the advantages from both F2F and F2M measurements are combined, we can obtain an accurate, drift-free, and tightly-coupled LVIO. The remaining problem is how to incorporate drift-free F2M measurement while reducing impacts on the LVIO state estimator.

The F2M constraint, as an absolute constraint, is very similar to the loop-closure constraint. Hence, integrating F2M measurements in a separate estimator is an effective solution. For example, the LiDAR F2M-based loop closure, LiDAR odometry, IMU preintegration, and global positioning system (GPS) measurements are fused in a pose-graph optimizer in LIO-SAM [28]. In R2LIVE [18], the LiDAR poses derived from the F2M-based LIO are incorporated into a sliding-window VIO to improve the accuracy of visual landmarks. However, the inconsistent problem caused by the F2M measurement must be solved when the state estimation is conducted in a unified estimator with both LiDAR, visual, and IMU measurements. If we can integrate the F2M absolute constraints while without affecting the covariance matrix, accurate state estimation can be obtained.

### *C. FGO or MSCKF*

Visual and LiDAR feature tracking are commonly conducted to derive F2F measurements across multiple frames. As a result, the FGO and MSCKF are typically adopted to incorporate these multi-state F2F measurements [2], [13], [14], [29], [31]. Besides, FGO has been proven more accurate than MSCKF through multiple linearization, especially when linearization errors are very large [27]. However, both the FGO and MSCKF involve a large amount of measurements from multiple frames, increasing the computational complexity. LIC-Fusion 2.0 [23] and VILENS-LVI [25] almost fail to run in real-time, even on desktop computers. Hence, how to improve the state-estimation efficiency with both LiDAR and visual measurements from multiple frames is very challenging.

In terms of F2F-based LIO, the inefficient estimation is mainly caused by the large number of separated point-to-plane or point-to-line models, *e.g.*, FF-LINS [11] and LIO-Mapping [32]. BA-LINS [13] employs an F2F plane-point bundle-adjustment (BA) model with an MSC form, and its efficiency is improved by more than 28%. As for F2F-based visual measurements, landmark states such as the inverse-depth parameter [33] are incorporated into the state vector for online estimation in conventional bundle adjustment (BA). However, the size of landmark states is almost the same as the size of pose states, and large computational resources are consumed to estimate visual landmarks [14]. A pose-only [34], [35], [36] visual-inertial state estimator is proposed in PO-VINS without involving landmarks in the state vector, and it exhibits 33% and 56% improved efficiency on a laptop and an onboard computer [14], respectively. Hence, we can derive an efficient and F2F-based LVIO by integrating the LiDAR BA and visual pose-only measurement, even within the FGO framework.

### D. *Our Contributions*

In this study, we propose a novel LVIO with a pose-only bundle adjustment (PA) framework, named PA-LVIO, to meet the urgent need for real-time navigation and mapping. Here, the "pose-only" denotes that no additional landmark state is involved, while the "bundle adjustment" denotes that the LiDAR and visual measurement models have an MSC form.

PA-LVIO is robust and accurate in complex scenarios by tightly integrating LiDAR, visual, and IMU measurements. Benefiting from the F2F LiDAR and visual PA measurement models, the proposed PA-LVIO is highly efficient, even using the nonlinear state estimator. Meanwhile, we employ a marginalization-free LiDAR F2M constraint to eliminate the drifts of LVIO, while reducing the impacts on the state estimator. Besides, we adopt an IMU-centric spatial-temporal calibration to achieve a pixel-wise alignment for LiDAR and camera, and thus a RGB-rendered point-cloud map can be obtained. The main contributions of our work are as follows:

- An accurate, efficient, and tightly-coupled LVIO, named PA-LVIO, with a novel pose-only BA for both LiDAR and visual measurements, is presented, to satisfy the urgent demand for real-time odometry and mapping.
- A marginalization-free LiDAR F2M measurement model is integrated to eliminate odometry drifts while reducing the influences on the estimator. Specifically, the LiDAR pose derived from a F2M optimization is employed for higher state-estimation efficiency.
- An IMU-centric online spatial-temporal calibration and compensation is employed to improve the robustness and accuracy. Besides, a pixel-wise LiDAR-camera alignment can be obtained from the online calibration, and thus an RGB-rendered point-cloud map can be built.
- Exhaustive experiments are conducted on both public and private datasets collected by wheeled robot, UAV, and handheld devices with a total length of more than 50 km. The results demonstrate the superior performance of the proposed PA-LVIO regarding accuracy, efficiency, and mapping quality.

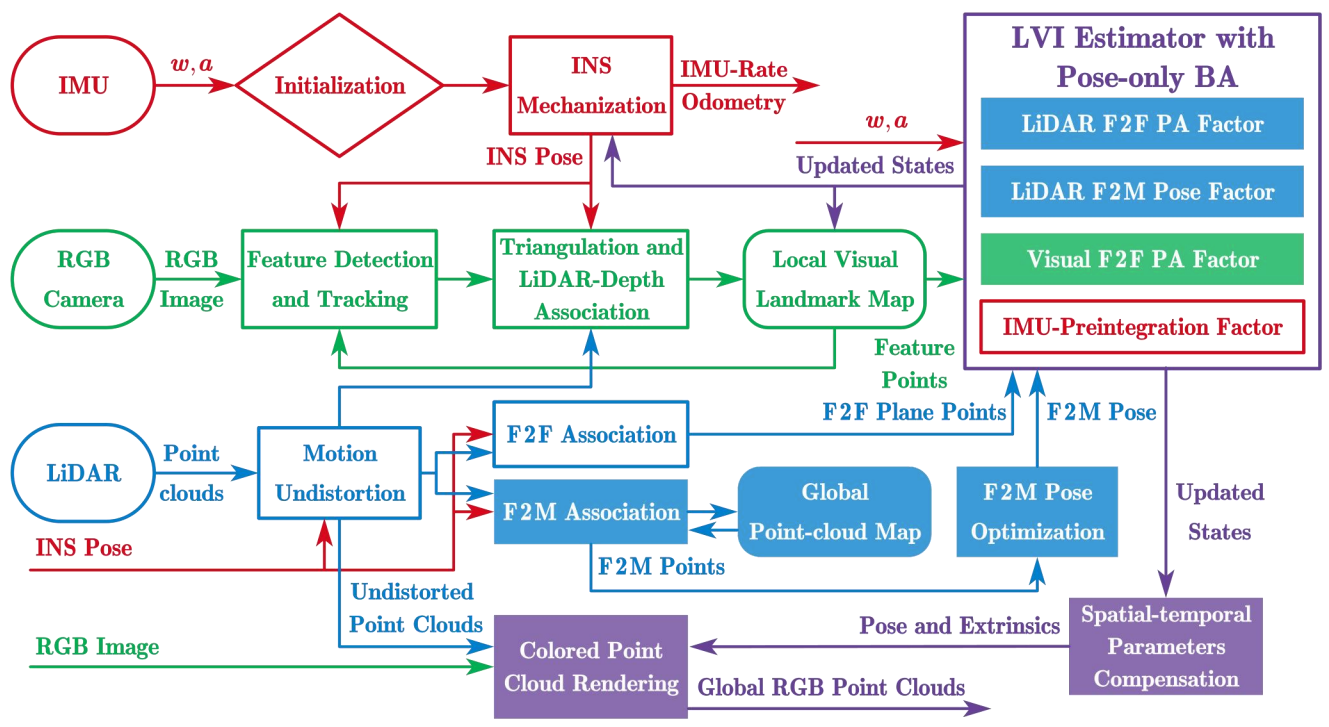

Fig. 1. System overview of the proposed PA-LVIO. The filled blocks denote the works in this paper.

## II. System Overview

The pipeline of the proposed PA-LVIO is exhibited in Fig. 1. We adopt an INS-centric framework to process data from the 3D LiDAR and RGB camera. The prior pose from INS are employed to assist both the LiDAR and visual data associations, which have been proven robust and efficient in current works [2], [11], [13], [14], [37]. The LiDAR and visual measurements together with the IMU measurements are tightly integrated under the framework of sliding-window FGO. A fixed size of LiDAR and visual keyframes are maintained in the sliding window to achieve an accurate and efficient state estimation.

With the high-rate INS pose, visual feature points are detected and tracked frame by frame. Visual keyframes are selected according to keyframe intervals and relative motions. Triangulation and LiDAR-depth association are conducted to obtain depths for visual landmarks. The visual landmarks with their feature points are inserted into a local map for efficient management. Here, the landmarks depths are only used for outlier culling of feature points [2], [14], as they are not contained in the state estimator.

At the same time, point clouds from LiDAR are undistorted using the high-rate INS pose. The LiDAR frames are projected to the time of visual keyframes to facilitate the data process. As a result, the LiDAR keyframe and the visual keyframe are synchronized to the same time. We follow the same method in BA-LINS [13] to associate same-plane points across multiple frames. Meanwhile, we adopt a F2M method to associate the latest and oldest keyframes with the global point-cloud map, which is managed with the ikd-Tree [12]. A F2M pose optimization is conducted to obtain the absolute pose relative to the global map. Hence, only the F2M pose is employed in the LVIO estimator to improve efficiency.

The LiDAR, visual, and IMU measurements are fused using a sliding-window optimizer. Specifically, the LiDAR plane points are employed to construct the F2F PA factor with an MSC form to provide relative pose constraints. The visual feature points are used as the F2F PA factor without involving landmark states. The IMU measurements are adopted as the IMU-preintegration factor [3]. Besides, the LiDAR F2M pose is adopted to construct the F2M pose factor, which provides absolute constraints to the global map and thus eliminates drifts. Meanwhile, the LiDAR-IMU and camera-IMU spatial and temporal parameters are all incorporated into the estimator for online calibration and compensation.

Once the state estimation is finished, the oldest keyframe in the sliding window together with all related measurements are marginalized to construct the prior information [38]. However, the LiDAR F2M pose measurement corresponding to the oldest keyframe is discarded directly without marginalization to reduce the negative impacts on the estimator. With the estimated pose and extrinsics, we can conduct a pixel-wise LiDAR-camera alignment and build a RGB-rendered point-cloud map using the marginalized LiDAR point clouds and the visual image.

## III. METHODOLOGY

### A. Problem Formulation

The proposed tightly-coupled LiDAR-visual-inertial state estimator PA-LVIO is a sliding-window optimizer. In particular, the LiDAR and visual F2F measurements are in a pose-only bundle-adjustment (PA) form for higher accuracy and efficiency. Hence, no landmark state is included in the state vector $\boldsymbol{X}$. Besides, the PA measurements have a MSC form and provide relative pose constraints between multiple frames.

The state vector $\boldsymbol{X}$ in the sliding window is defined as follows

$$\begin{aligned} \boldsymbol{X} &= \left[\boldsymbol{x}_0, \boldsymbol{x}_1, \ldots, \boldsymbol{x}_n, \boldsymbol{x}_{\mathrm{c}}^{\mathrm{b}}, \boldsymbol{x}_{\mathrm{r}}^{\mathrm{b}}\right] \\ \boldsymbol{x}_k &= \left[\boldsymbol{p}_{\mathrm{wb}_k}^{\mathrm{w}}, \mathbf{q}_{\mathrm{b}_k}^{\mathrm{w}}, \boldsymbol{v}_{\mathrm{wb}_k}^{\mathrm{w}}, \boldsymbol{b}_{g_k}, \boldsymbol{b}_{a_k}, \delta t_{\mathrm{bc},k}\right], k \in [0, n] \\ \boldsymbol{x}_{\mathrm{c}}^{\mathrm{b}} &= \left[\boldsymbol{p}_{\mathrm{bc}}^{\mathrm{b}}, \mathbf{q}_{\mathrm{c}}^{\mathrm{b}}\right], \boldsymbol{x}_{\mathrm{r}}^{\mathrm{b}} = \left[\boldsymbol{p}_{\mathrm{br}}^{\mathrm{b}}, \mathbf{q}_{\mathrm{r}}^{\mathrm{b}}, \delta t_{\mathrm{br}}\right] \end{aligned} \quad (1)$$

where $\boldsymbol{x}_k$ is the state vector at each keyframe, including the IMU (body frame, b-frame) position $\boldsymbol{p}_{\mathrm{wb}_k}^{\mathrm{w}}$, attitude quaternion $\mathbf{q}_{\mathrm{b}_k}^{\mathrm{w}}$, and velocity $\boldsymbol{v}_{\mathrm{wb}_k}^{\mathrm{w}}$ in the world frame (w-frame). $\boldsymbol{b}_g$ and $\boldsymbol{b}_a$ are the gyroscope and accelerometer biases, respectively. $\delta t_{\mathrm{bc},k}$ denotes the time-varying time-delay parameter between IMU and camera (camera-frame, c-frame). $\boldsymbol{x}_{\mathrm{c}}^{\mathrm{b}}$ and $\boldsymbol{x}_{\mathrm{r}}^{\mathrm{b}}$ are the camera-IMU and LiDAR-camera extrinsic parameters, respectively. $\delta t_{\mathrm{br}}$ is the time-delay parameter between IMU and LiDAR (LiDAR-frame, r-frame). $n$ is the size of the sliding window.

As the visual landmark state parameter, such as the depth or the 3D position, is not contained in (1), the proposed PA-LVIO is a pose-only form regarding LiDAR and visual measurements. The FGO-based state estimation in PA-LVIO is achieved by minimizing the sum of the Mahalanobis norm [39] of all measurements and the prior as

$$\underset{\boldsymbol{X}}{\operatorname{argmin}} \frac{1}{2} \left\{ \begin{aligned} &\sum_{i \in [1,h]} \left\| \boldsymbol{e}^R\left(\tilde{\mathbf{z}}^{R_i}, \boldsymbol{X}\right) \right\|_{\boldsymbol{\Sigma}^{R_i}}^2 + \sum_{i \in \{0,n\}} \left\| \boldsymbol{e}^L\left(\tilde{\mathbf{z}}^{L_i}, \boldsymbol{X}\right) \right\|_{\boldsymbol{\Sigma}^{L_i}}^2 \\ &+ \sum_{j \neq \varsigma, i \in [1,m]} \left\| \boldsymbol{e}^V\left(\tilde{\mathbf{z}}_{j,l_i}^{V_{\varsigma,\eta}}, \boldsymbol{X}\right) \right\|_{\boldsymbol{\Sigma}_{l_i}^{V_{\varsigma,\eta}}}^2 \\ &+ \sum_{i \in [1,s]} \left\| \boldsymbol{e}^D\left(\tilde{\mathbf{z}}^{D_i}, \boldsymbol{X}\right) \right\|_{\boldsymbol{\Sigma}^{D_i}}^2 \\ &+ \sum_{k \in [1,n]} \left\| \boldsymbol{e}^T\left(\tilde{\mathbf{z}}_{k-1,k}^T, \boldsymbol{X}\right) \right\|_{\boldsymbol{\Sigma}_{k-1,k}^T}^2 \\ &+ \sum_{k \in [1,n]} \left\| \boldsymbol{e}^I\left(\tilde{\mathbf{z}}_{k-1,k}^I, \boldsymbol{X}\right) \right\|_{\boldsymbol{\Sigma}_{k-1,k}^I}^2 + \left\| \boldsymbol{e}^M\left(\tilde{\mathbf{z}}^M, \boldsymbol{X}\right) \right\|^2 \end{aligned} \right\} \quad (2)$$

where $\boldsymbol{e}^R$ is the residual of the LiDAR PA factor [13]; $\boldsymbol{e}^L$ is the residual of the LiDAR F2M pose factor; $\boldsymbol{e}^V$ is the residual of the visual PA factor [14]; $\boldsymbol{e}^D$ is the residual of the LiDAR-depth PA factor [14], which combines accurate LiDAR depth and visual feature measurements; $\boldsymbol{e}^T$ is the residual of the camera-IMU time-delay factor, which provides relative constraints for the time-varying time-delay parameter; $\boldsymbol{e}^I$ is the residual of the IMU-preintegration factor [3]; $\boldsymbol{e}^M$ is the prior residual, which is derived from the marginalization [38]; $\boldsymbol{\Sigma}$ is the covariance for each factor. $h$, $s$, and $m$ denote the total LiDAR same-plane point, LiDAR-depth, and visual landmark measurements, respectively.

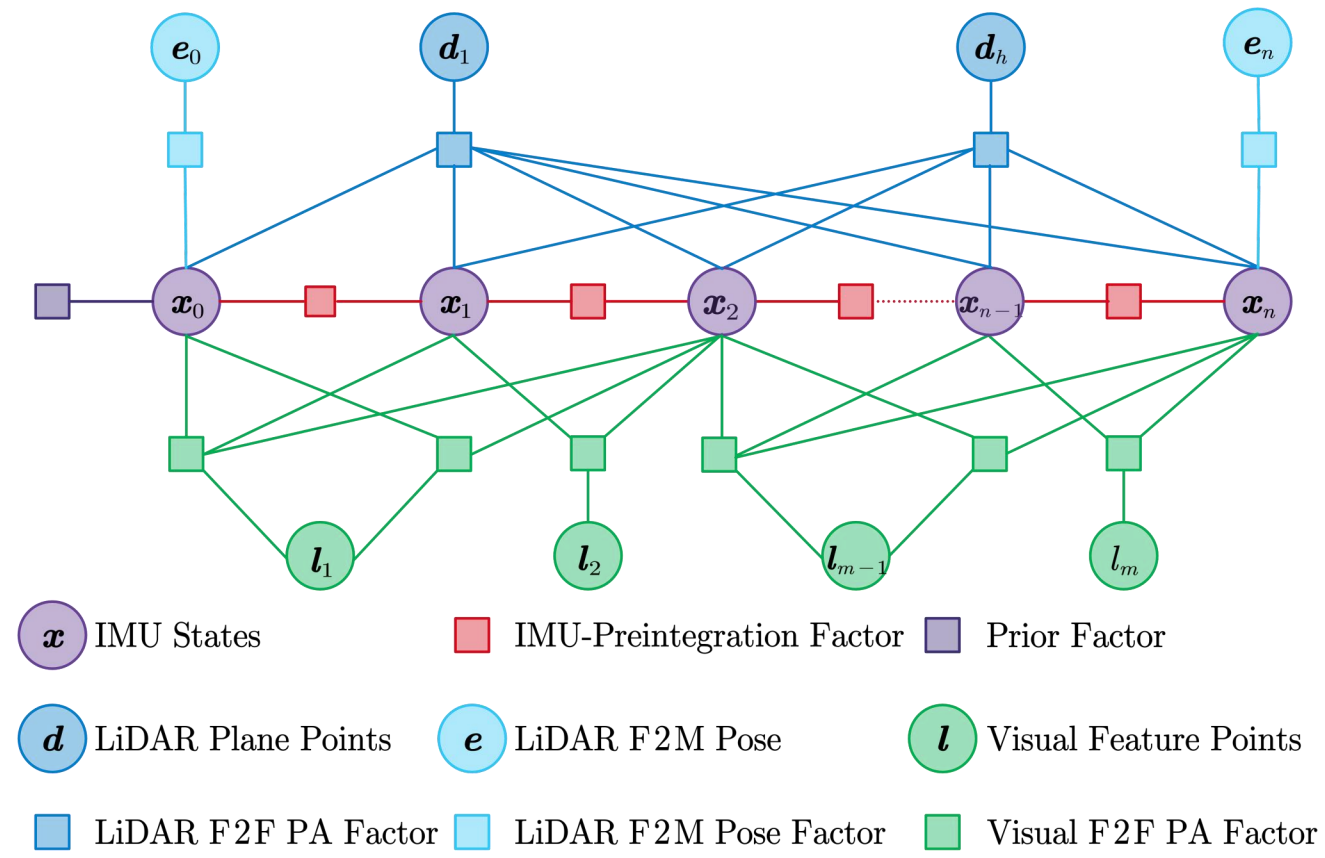

Fig. 2. FGO framework of the proposed PA-LVIO. The LiDAR-depth PA factor, which combines accurate LiDAR depth and visual feature measurements, are not exhibited for better visualization.

Fig. 2 shows the FGO framework of the proposed PA-LVIO. The LiDAR and visual F2F factors provide relative pose constraints between multiple frames without involving landmark states, and thus they are in a PA form. Besides, the LiDAR F2M pose factor only provides absolute pose constraints to the latest and oldest IMU pose in the sliding window, to reduce computations. Ceres Solver [40], an open-sourced library, is adopted to solve the nonlinear optimization problem in (2) using the Levenberg-Marquardt (LM) algorithm. Besides, the Huber robust cost function [39] is employed to reduce the impacts of outliers.

### B. Visual F2F PA Factor

In the proposed PA-LVIO, visual landmarks can be classified into two categories, *i.e.*, with or without LiDAR

depth. Hence, the visual factors include feature reprojection factor and LiDAR-depth factor, which are all in F2F and PA forms and provide relative pose constraints across multiple frames, as exhibited in Fig. 2. The former includes only feature observations, which the latter includes both feature and LiDAR depth observations.

For a visual landmark $l_i$, its two anchored keyframes are expressed as $\mathbf{F}_\varsigma$ and $\mathbf{F}_\eta$. The anchored frame $\mathbf{F}_\varsigma$ is the first observed keyframe or the keyframe associated with LiDAR depth, while $\mathbf{F}_\eta$ is selected by maximizing the parallax [14]. According to the pose-only multiply-view constraints [14], [34], the landmark depth in $\mathbf{F}_\varsigma$ can be expressed as a function of the camera pose of the anchored keyframes as

$$d_\varsigma^{(\varsigma,\eta)} = \frac{\left\| [\tilde{\boldsymbol{p}}^{\mathrm{u}_\eta}]_\times \boldsymbol{p}_{\mathrm{c}_\eta \mathrm{c}_\varsigma}^{\mathrm{c}_\eta} \right\|}{\theta_{\varsigma,\eta}} \tag{3}$$

where $\tilde{\boldsymbol{p}}^{\mathrm{u}}$ is the coordinate of the feature point in the normalized camera frame (u-frame) [14], and the parallax parameter $\theta_{\varsigma,\eta}$ can be written as

$$\theta_{\varsigma,\eta} = \left\| [\tilde{\boldsymbol{p}}^{\mathrm{u}_\eta}]_\times \mathbf{R}_{\mathrm{c}_\varsigma}^{\mathrm{c}_\eta} \tilde{\boldsymbol{p}}^{\mathrm{u}_\varsigma} \right\| \tag{4}$$

The relative camera pose $\left\{ \boldsymbol{p}_{\mathrm{c}_\eta \mathrm{c}_\varsigma}^{\mathrm{c}_\eta}, \mathbf{R}_{\mathrm{c}_\varsigma}^{\mathrm{c}_\eta} \right\}$ can be calculated using the IMU pose and camera-IMU extrinsic states in (1) [14] as

$$\begin{cases} \boldsymbol{p}_{\mathrm{wc}}^{\mathrm{w}} = \boldsymbol{p}_{\mathrm{wb}}^{\mathrm{w}} + \mathbf{R}_{\mathrm{b}}^{\mathrm{w}} \boldsymbol{p}_{\mathrm{bc}}^{\mathrm{b}} \\ \mathbf{R}_{\mathrm{c}}^{\mathrm{w}} = \mathbf{R}_{\mathrm{b}}^{\mathrm{w}} \mathbf{R}_{\mathrm{c}}^{\mathrm{b}} \end{cases} \tag{5}$$

$$\begin{cases} \boldsymbol{p}_{\mathrm{c}_\eta \mathrm{c}_\varsigma}^{\mathrm{c}_\eta} = \left( \mathbf{R}_{\mathrm{c}_\eta}^{\mathrm{w}} \right)^{-1} \left( \boldsymbol{p}_{\mathrm{wc}_\varsigma}^{\mathrm{w}} - \boldsymbol{p}_{\mathrm{wc}_\eta}^{\mathrm{w}} \right) \\ \mathbf{R}_{\mathrm{c}_\varsigma}^{\mathrm{c}_\eta} = \left( \mathbf{R}_{\mathrm{c}_\eta}^{\mathrm{w}} \right)^{-1} \mathbf{R}_{\mathrm{c}_\varsigma}^{\mathrm{w}} \end{cases} \tag{6}$$

For a visual landmark $l_i$ without LiDAR depth, one of its observed keyframes is $\mathbf{F}_j, j \neq \varsigma$. The visual PA residual can be written as

$$\boldsymbol{e}^V \left( \tilde{\boldsymbol{z}}_{j,l_i}^{V_{\varsigma,\eta}}, \boldsymbol{X} \right) = [\boldsymbol{b}_1 \quad \boldsymbol{b}_2]^T \left( \frac{\hat{\boldsymbol{p}}^{\mathrm{c}_j}}{\|\hat{\boldsymbol{p}}^{\mathrm{c}_j}\|} - \tilde{\boldsymbol{p}}^{\mathrm{u}_j} \right) \tag{7}$$

where $\hat{\boldsymbol{p}}^{\mathrm{c}_j} = d_\varsigma^{(\varsigma,\eta)} \mathbf{R}_{\mathrm{c}_\varsigma}^{\mathrm{c}_j} \tilde{\boldsymbol{p}}^{\mathrm{u}_\varsigma} + \boldsymbol{p}_{\mathrm{c}_j \mathrm{c}_\varsigma}^{\mathrm{c}_j}$ is the calculated coordinate in the c-frame of the keyframe $\mathbf{F}_j$. $\boldsymbol{b}_1 = [1 \quad 0 \quad 0]^T$ and $\boldsymbol{b}_2 = [0 \quad 1 \quad 0]^T$ are two orthogonal bases. The covariance matrix $\boldsymbol{\Sigma}_{l_i}^{V_{\varsigma,\eta}}$ is propagated from the pixel plane onto the tangent plane. If the visual landmark is associated with LiDAR depth, we employ an MSC-based LiDAR-depth factor, which is also in the PA form [14]. Specifically, we compress all feature observations of the landmark and its LiDAR depth into a single measurement while maintaining the pose-only form. We can refer to PO-VINS [14] to derive the LiDAR-depth PA residual $\boldsymbol{e}^D(\tilde{\boldsymbol{z}}^{D_i}, \boldsymbol{X})$.

### C. LiDAR F2F PA Factor

We adopt a plane-point BA factor [13], [29], which is F2F and PA form and provides relative pose constraints across multiple frames, as shown in Fig. 2. The LiDAR PA residual $\boldsymbol{e}^R(\tilde{\boldsymbol{z}}^{R_i}, \boldsymbol{X})$ is equivalent to the plane thickness of the associated same-plane points, which are from multiple frames. For a cluster of same-plane points in the LiDAR frame (r-frame) $\tilde{\boldsymbol{p}}^{\mathrm{r}_j}, j \in \mathbb{C}_i$, where $\mathbb{C}_i$ denotes the keyframe collections of the points, they can be projected to the w-frame using IMU pose and LiDAR-IMU extrinsics as

$$\hat{\boldsymbol{p}}^{\mathrm{w}} = \mathbf{R}_{\mathrm{b}}^{\mathrm{w}} \left( \mathbf{R}_{\mathrm{r}}^{\mathrm{b}} \tilde{\boldsymbol{p}}^{\mathrm{r}} + \boldsymbol{p}_{\mathrm{br}}^{\mathrm{b}} \right) + \boldsymbol{p}_{\mathrm{wb}}^{\mathrm{w}} \tag{8}$$

We can then obtain the plane parameters $\{\boldsymbol{n}, d\}$ by solving an over-determined linear equation. A point-to-plane distance $\varepsilon$ for a point $\hat{\boldsymbol{p}}^{\mathrm{w}}$ to the plane is written as

$$\varepsilon = \boldsymbol{n}^T \hat{\boldsymbol{p}}^{\mathrm{w}} + d \tag{9}$$

The plane thickness can be defined as the average square point-to-plane distance for all points [13]. Hence, the LiDAR PA residual $\boldsymbol{e}^R(\tilde{\boldsymbol{z}}^{R_i}, \boldsymbol{X})$ in (2) can be calculated as

$$\boldsymbol{e}^R(\tilde{\boldsymbol{z}}^{R_i}, \boldsymbol{X}) = \frac{1}{N} \sum_{j \in \mathbb{C}_i}^{N} (\boldsymbol{n}^T \hat{\boldsymbol{p}}^{\mathrm{w}_j} + d)^2 \tag{10}$$

The covariance matrix $\boldsymbol{\Sigma}^{R_i}$ is adaptively estimated by quantitative statistics of the plane thicknesses using points from different keyframes [13].

### D. Marginalization-Free F2M Pose Factor

We can obtain an accurate LVIO by tightly integrating the visual and LiDAR F2F PA factors together with the IMU-preintegration factors, which all provide relative pose constraints. However, the odometry may drift over time or distance without using other absolute pose constraints, even in small-scale scenarios. Hence, we propose a LiDAR marginalization-free F2M pose factor. The F2M pose constraints are integrated to eliminate drifts, while they are not marginalized to reduce the impacts on the state estimator. Here, the F2M pose rather than raw measurements are adopted to reduce computational costs.

Specifically, we associate the latest keyframe $\mathbf{F}_n$ and the oldest keyframe $\mathbf{F}_0$ with the global point-cloud map, which is similar to [12]. The point clouds from the marginalized LiDAR keyframe will be added into the global point-cloud map for further associations. For a point $\tilde{\boldsymbol{p}}^{\mathrm{r}_j}$ in the keyframe $\mathbf{F}_i, i \in \{0, n\}$, we can obtain the corresponding plane parameters $\{\boldsymbol{n}_j, d_j\}$ using the associated points in the global map. The F2M pose $\{\boldsymbol{p}_{\mathrm{wr}_i}^{\mathrm{w}}, \mathbf{q}_{\mathrm{r}_i}^{\mathrm{w}}\}$ can be estimated by minimizing the point-to-plane distances of all associated points as

$$\arg \min_{\{\boldsymbol{p}_{\mathrm{wr}_i}^{\mathrm{w}}, \mathbf{q}_{\mathrm{r}_i}^{\mathrm{w}}\}} \frac{1}{2} \sum_j \left\| \boldsymbol{n}_j^T \left( \mathbf{R}_{\mathrm{r}_i}^{\mathrm{w}} \tilde{\boldsymbol{p}}^{\mathrm{r}_j} + \boldsymbol{p}_{\mathrm{wr}_i}^{\mathrm{w}} \right) + d_j \right\|^2 \tag{11}$$

We can also obtain the covariance matrix $\boldsymbol{\Sigma}^L$ of the estimated pose. Hence, the F2M pose residual in (2) can be expressed as

$$\boldsymbol{e}^L(\tilde{\boldsymbol{z}}^{L_i}, \boldsymbol{X}) = \begin{bmatrix} \mathbf{R}_{\mathrm{b}_i}^{\mathrm{w}} \boldsymbol{p}_{\mathrm{br}}^{\mathrm{b}} + \boldsymbol{p}_{\mathrm{wb}_i}^{\mathrm{w}} - \boldsymbol{p}_{\mathrm{wr}_i}^{\mathrm{w}} \\ \mathrm{Log} \left( \mathbf{q}_{\mathrm{b}_i}^{\mathrm{w}} \otimes \mathbf{q}_{\mathrm{r}}^{\mathrm{b}} \otimes \left( \mathbf{q}_{\mathrm{r}_i}^{\mathrm{w}} \right)^{-1} \right) \end{bmatrix} \tag{12}$$

where $i \in \{0, n\}$ denotes the keyframe $\mathbf{F}_0$ or $\mathbf{F}_n$. As only two F2M pose factors are included in (2), the proposed state estimator is efficient.

When the nonlinear optimization problem in (2) is solved, the marginalization will be conducted to convert the oldest keyframe with related measurements into prior information. However, the F2M pose $\{\boldsymbol{p}_{\mathrm{wr}_0}^{\mathrm{w}}, \mathbf{q}_{\mathrm{r}_0}^{\mathrm{w}}\}$ are discarded directly without marginalization to reduce the effects on the estimator. Fig. 3 depicts the estimated standard deviation (STD) of the yaw angle using different methods. If the F2M measurement is marginalized, the yaw STD will not diverge, demonstrating inconsistent estimation [11]. In contrast, the proposed

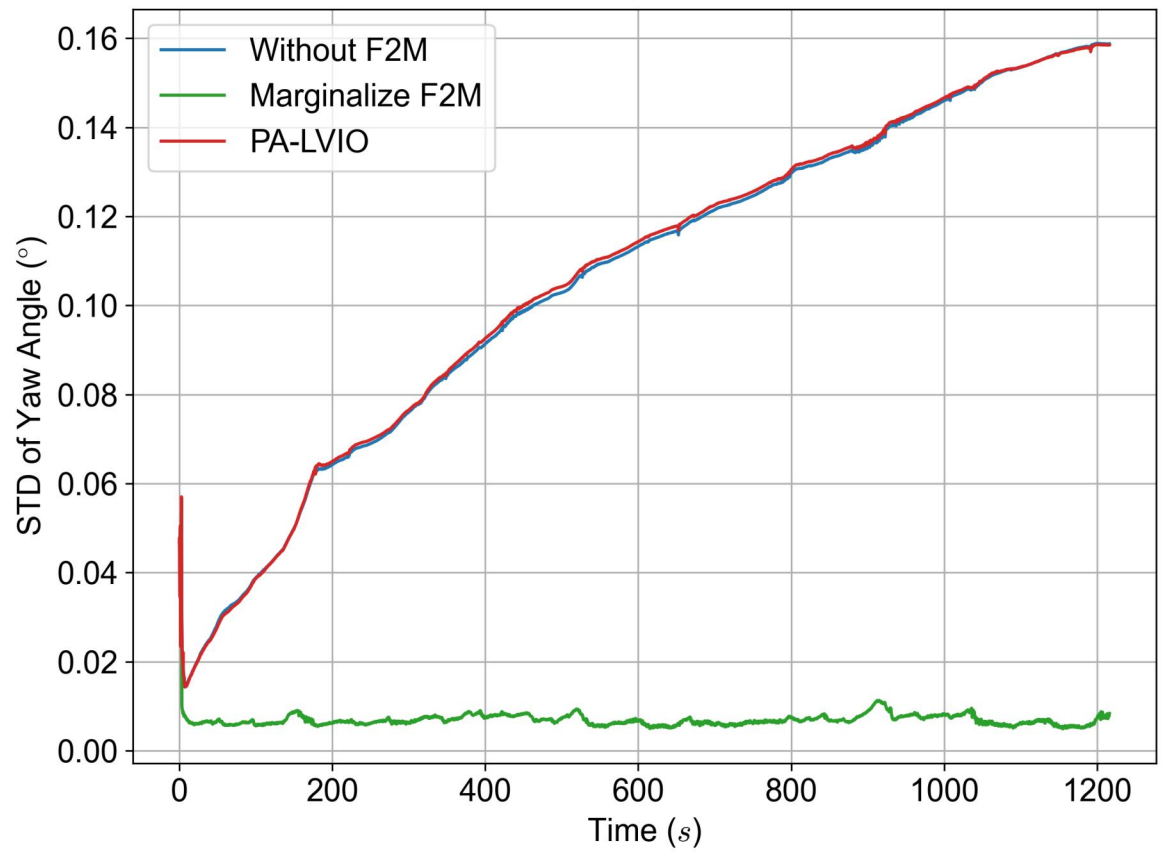

Fig. 3. Comparison of the estimated standard deviation (STD) of the yaw angle using different methods.

PA-LVIO exhibits similar results to the method without involving the F2M factor, which means more consistent estimation. It should be noted that the F2M absolute constraint may have greater returns in the long term, due to the gradually diverging covariance. Nevertheless, the state estimator is not optimal by incorporating the F2M absolute constraint, even without marginalization. As the odometry drifts can be eliminated, the proposed marginalization-free F2M pose factor is effective and applicable, especially for LVIO applications.

### E. IMU-Centric Spatial-Temporal Alignment

We employ an IMU-centric spatial-temporal alignment to online calibrate and compensate the camera-IMU and LiDAR-IMU parameters. The online calibration can improve the robustness and accuracy of PA-LVIO notably. Meanwhile, we can obtain a pixel-wise LiDAR-camera alignment, and thus can build a RGB-rendered point-cloud map. Both the intrinsic and time-delay parameters are included in the state vector (1) for online calibration. The extrinsic parameters have already been considered in residuals (7), (10), and (12), while the time-delay parameters are omitted. Besides, the extrinsics are modeled as random constants, including the translation and rotation parameters.

#### 1) Temporal Calibration for Camera and IMU

The camera-IMU time parameter $\delta t_{\mathrm{bc}}$ is modeled as a random walk process, as the exposure time of the camera is time-varying. $\delta t_{\mathrm{bc}}$ is defined as the time offset between the IMU timestamp $t_{\mathrm{b}}$ and the camera timestamp $t_{\mathrm{c}}$

$$t_{\mathrm{b}} = t_{\mathrm{c}} + \delta t_{\mathrm{bc}} \tag{13}$$

The continuous-time state equation for $\delta t_{\mathrm{bc}}$ is given by

$$\delta \dot{t}_{\mathrm{bc}} = w_{\delta t_{\mathrm{bc}}} \tag{14}$$

where $w_{\delta t_{\mathrm{bc}}}$ is the driving white noise. The calibrated time-delay parameter for keyframe $\mathbf{F}_k$ can be expressed as $\hat{\delta} t_{\mathrm{br},k}$ , and the bias $\Delta t_{\mathrm{bc},k}$ between $\hat{\delta} t_{\mathrm{br},k}$ and the estimating $\delta t_{\mathrm{br},k}$ can be written as

$$\Delta t_{\mathrm{bc},k} = \delta t_{\mathrm{bc},k} - \hat{\delta} t_{\mathrm{bc},k} \tag{15}$$

The bias $\Delta t_{\mathrm{bc},k}$ can be calibrated using a constant-velocity model on the pixel plane [41]. The feature velocity $\boldsymbol{v}^{\mathrm{u}_k}$ in the u-frame $\boldsymbol{p}^{\mathrm{u}_k}$ can be calculated using two continuously tracked feature coordinates as

$$\boldsymbol{v}^{\mathrm{u}_k} = \frac{\boldsymbol{p}^{\mathrm{u}_2} - \boldsymbol{p}^{\mathrm{u}_1}}{\Delta t_{12}} \tag{16}$$

where $\Delta t_{12}$ is the time interval, and $\boldsymbol{p}^{\mathrm{u}_1}$ is observed in the keyframe $\mathbf{F}_k$. The raw feature observation $\boldsymbol{p}^{\mathrm{u}_k}$ can be calibrated using a constant-velocity model to obtain the feature observation $\tilde{\boldsymbol{p}}^{\mathrm{u}_k}$ in Section III.B as

$$\tilde{\boldsymbol{p}}^{\mathrm{u}_k} = \boldsymbol{p}^{\mathrm{u}_k} - \boldsymbol{v}^{\mathrm{u}_k} \Delta t_{\mathrm{bc},k} \tag{17}$$

The calibrated feature $\tilde{\boldsymbol{p}}^{\mathrm{u}_k}$ can be then used to derive visual PA residual in (7), and thus $\delta t_{\mathrm{bc}}$ can be integrated into the estimator. As the time parameter $\delta t_{\mathrm{bc}}$ is modeled as a random walk process, the residual for the time-delay factor in (2) can be written as

$$\boldsymbol{e}^{T}\left(\tilde{\boldsymbol{z}}_{k-1,k}^{T}, \boldsymbol{X}\right) = \delta t_{\mathrm{bc},k} - \delta t_{\mathrm{bc},k-1} \tag{18}$$

The covariance matrix $\boldsymbol{\Sigma}_{k-1,k}^{T}$ can be propagated using the driving noise of the random walk process as

$$\boldsymbol{\Sigma}_{k-1,k}^{T} = \sigma_{t_{\mathrm{bc}}}^{2} \Delta t_{k-1,k} \mathbf{I} \tag{19}$$

where $\Delta t_{k-1,k}$ is the interval, and $\sigma_{t_{\mathrm{bc}}}^{2}$ is the covariance of $w_{\delta t_{\mathrm{bc}}}$.

#### 2) Temporal Calibration for LiDAR and IMU

The LiDAR-IMU time-delay parameter $\delta t_{\mathrm{br}}$ is modeled as a random constant, mainly because it will hardly change when LiDAR and IMU are well synchronized. $\delta t_{\mathrm{br}}$ is defined as the time offset between the IMU timestamp $t_{\mathrm{b}}$ and the LiDAR timestamp $t_{\mathrm{r}}$

$$t_{\mathrm{b}} = t_{\mathrm{r}} + \delta t_{\mathrm{br}} \tag{20}$$

Assuming the calibrated time-delay parameter is $\hat{\delta} t_{\mathrm{br},k}$ for keyframe $\mathbf{F}_k$, the time bias between $\hat{\delta} t_{\mathrm{br},k}$ and the estimating $\delta t_{\mathrm{br}}$ can be calculated as

$$\Delta t_{\mathrm{br},k} = \delta t_{\mathrm{br}} - \hat{\delta} t_{\mathrm{br},k} \tag{21}$$

$\Delta t_{\mathrm{br},k}$ can be calibrated using a constant-motion model. Specifically, we can obtain the velocity and angular velocity of the IMU at $\mathbf{F}_k$, denoted as $\tilde{\boldsymbol{v}}_{\mathrm{wb}_k}^{\mathrm{w}}$ and $\tilde{\boldsymbol{\omega}}_{\mathrm{wb}_k}^{\mathrm{w}}$. Hence, the IMU pose can be rewritten by considering the time bias $\Delta t_{\mathrm{br},k}$ as

$$\begin{cases} \hat{\boldsymbol{p}}_{\mathrm{wb}_k}^{\mathrm{w}} = \boldsymbol{p}_{\mathrm{wb}_k}^{\mathrm{w}} + \tilde{\boldsymbol{v}}_{\mathrm{wb}_k}^{\mathrm{w}} \Delta t_{\mathrm{br},k} \\ \hat{\mathbf{q}}_{\mathrm{b}_k}^{\mathrm{w}} = \mathbf{q}_{\mathrm{b}_k}^{\mathrm{w}} \otimes \mathrm{Exp}\left(\tilde{\boldsymbol{\omega}}_{\mathrm{wb}_k}^{\mathrm{w}} \Delta t_{\mathrm{br},k}\right) \end{cases} \tag{22}$$

Here, $\left\{\hat{\boldsymbol{p}}_{\mathrm{wb}_k}^{\mathrm{w}}, \hat{\mathbf{q}}_{\mathrm{b}_k}^{\mathrm{w}}\right\}$ is the calibrated IMU pose, and it can be employed in (10) and (12). Hence, $\delta t_{\mathrm{br}}$ can be integrated into the estimator for online calibration and compensation.

#### 3) LiDAR-Camera Alignment

Once the FGO problem is solved, we can obtain the estimated camera-IMU and LiDAR-IMU parameters, including spatial and temporal alignment parameters. The temporal offset for LiDAR and camera can be calibrated by projecting the LiDAR point clouds using high-rate IMU pose. The LiDAR-camera extrinsics $\{\boldsymbol{p}_{\mathrm{cr}}^{\mathrm{c}}, \mathbf{q}_{\mathrm{r}}^{\mathrm{c}}\}$ can be transformed as

$$\begin{cases} \boldsymbol{p}_{\mathrm{cr}}^{\mathrm{c}} = \left(\mathbf{R}_{\mathrm{c}}^{\mathrm{b}}\right)^{-1} \left(\boldsymbol{p}_{\mathrm{br}}^{\mathrm{b}} - \boldsymbol{p}_{\mathrm{bc}}^{\mathrm{b}}\right) \\ \mathbf{q}_{\mathrm{r}}^{\mathrm{c}} = \left(\mathbf{q}_{\mathrm{c}}^{\mathrm{b}}\right)^{-1} \mathbf{q}_{\mathrm{r}}^{\mathrm{b}} \end{cases} \tag{23}$$

We can achieve a pixel-wise LiDAR-camera alignment using the IMU-centric calibration, though we do not calibrate the LiDAR and camera directly. Experiment results are shown in Section IV.C.1 to demonstrate the accuracy of LiDAR-camera alignment.

With the estimated IMU pose and extrinsics, we can build a global colored point-cloud map. The LiDAR point clouds of the marginalized keyframe can be projected into the world frame using pose and LiDAR-IMU extrinsics. The LiDAR point clouds are rendered with the RGB values from the visual images, which is similar to the method in FAST-LIVO2 [7]. Specifically, the LiDAR point clouds are transformed to a pixel plane using LiDAR-camera extrinsics, and the color can be obtained by interpolating the RGB values of the neighborhood pixels. As a result, we can obtain a RGB-rendered point-cloud map in the global world frame.

## IV. Experiments and Results

### A. Datasets and Implementation

Comprehensive experiments are carried out on both public and private datasets, including wheeled robot, UAV, and handheld datasets, to fully examine the proposed PA-LVIO. Specifically, the public *i2Nav-Robot* [42] dataset is collected by a low-speed wheeled robot, and contains 10 sequences with a total length of 17.1 km. The public *MARS-LVIG* [43] dataset is an UAV dataset with higher speed, and 9 sequences with a length of 32.6 km are adopted. *R3LIVE* [4] is a public handheld dataset with 6 sequences involved. The above public datasets are employed to evaluate the odometry and mapping results of the proposed PA-LVIO. The private *HandNav* dataset with 3 sequences is adopted to evaluate the mapping quality and also the real-time performance of PA-LVIO. Fig. 4 shows the equipment employed in the *HandNav* dataset. Table I shows the detailed information of these datasets, and different type of sensors are involved for the evaluation.

The proposed PA-LVIO is implemented with C++ under the framework of the robot operation system (ROS). Multi-threading technology is employed in PA-LVIO for higher processing efficiency. Besides, the visual processes are accelerated by OpenCV with NVIDIA CUDA support. The odometry and mapping results are evaluated on a desktop PC (AMD Ryzen 9 9950X CPU and NVIDIA RTX 5090 GPU). The efficiency results on the *HandNav* dataset are conducted on an onboard ARM computer, *i.e.*, the NVIDIA Orin NX (6-core CPU with 8-GB RAM).

TABLE I
Datasets Descriptions

| Dataset | *i2Nav-Robot* | *MARS-LVIG* | *R3LIVE* | *HandNav* |
|---|---|---|---|---|
| Carrier | Wheeled robot | UAV | Handheld | Handheld |
| LiDAR | Hesai AT128, 10 Hz | Livox AVIA, 10 Hz | Livox AVIA, 10 Hz | Livox AVIA, 10 Hz |
| Camera | 1600*1200, 10 Hz | 2448*2048, 10 Hz | 1280*1024, 15 Hz | 1800*1200, 10 Hz |
| IMU | ADIS16465, 200 Hz | BMI088, 200 Hz | BMI088, 200 Hz | BMI088, 200 Hz |
| Sequences | 10 sequences, 17.1 km | 9 sequence, 32.6 km | 6 sequence, 5.0 km | 3 sequence, 1.5 km |

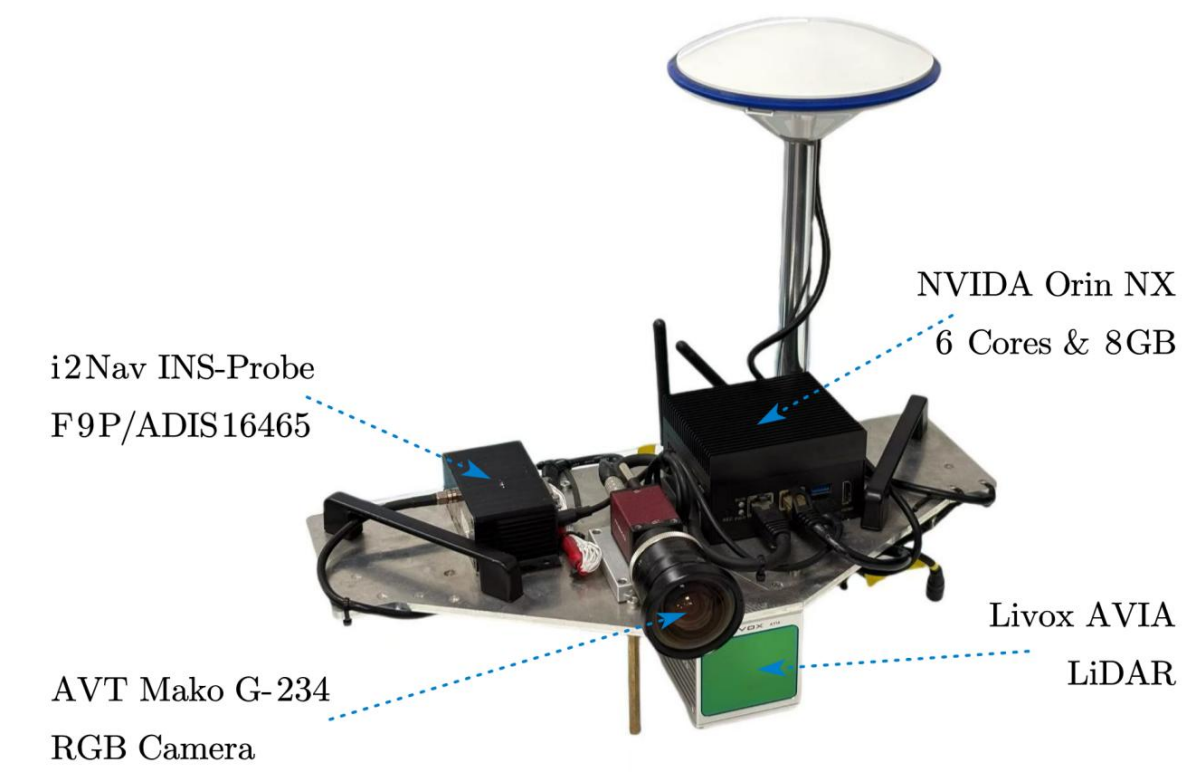

Fig. 4. Equipment setup in the *HandNav* dataset.

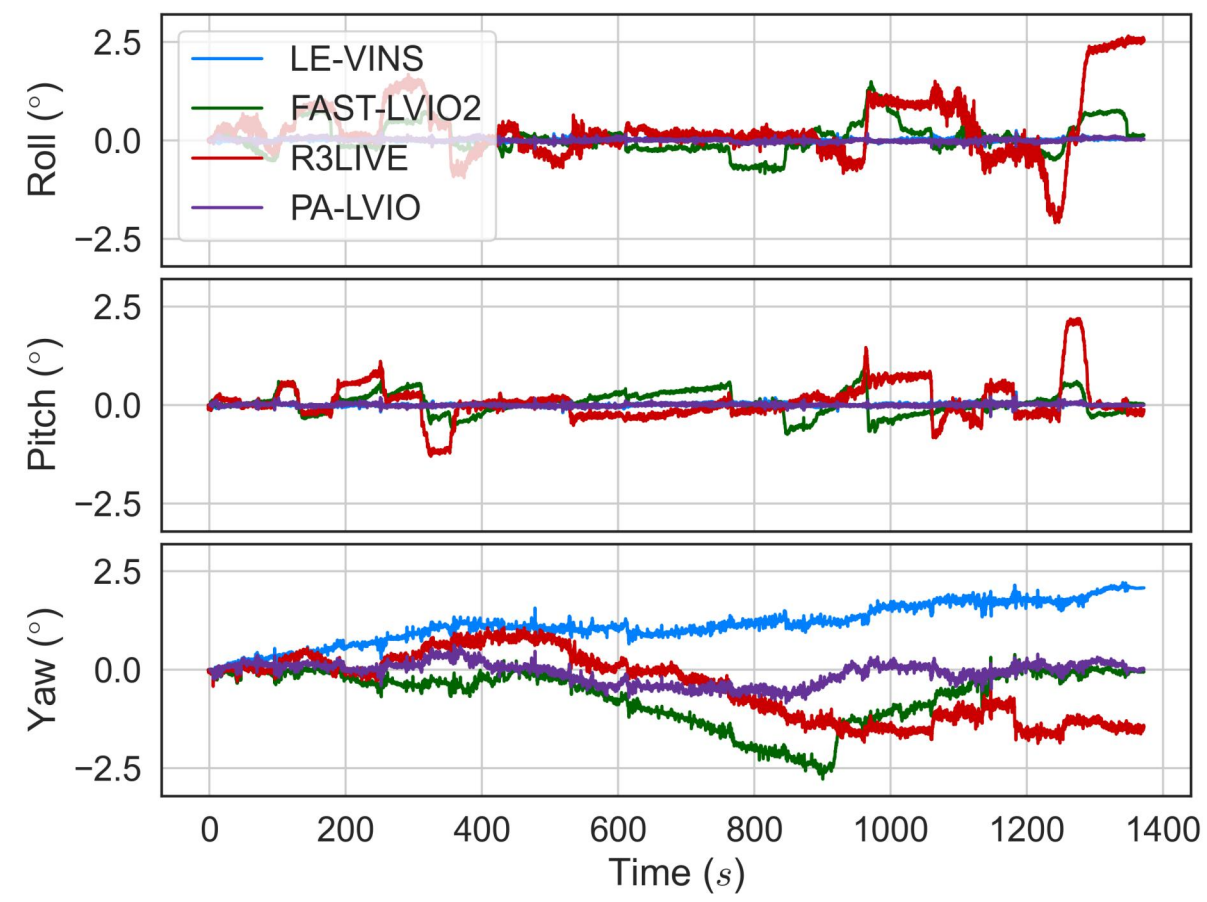

Fig. 5. The attitude errors on the *i2Nav-Robot-street00*. Here, only LiDAR-visual-inertial methods are involved for better visualization.

### B. Evaluation of Odometry

State-of-the-art (SOTA) odometry methods are employed for the evaluation, including FF-LINS [11], FAST-LIO2 [12], LE-VINS [2], R3LIVE [4], and FAST-LIVO2 [7]. FF-LINS is an F2F LIO using FGO, while FAST-LIO2 is F2M LIO using Kalman filter. LE-VINS is an FGO-based VIO with LiDAR-depth enhancement. These LIO and VIO methods are adopted for evaluating the subsystems of PA-LVIO. Here, the LIO subsystem of PA-LVIO includes the marginalization-free F2M pose factor, as presented in Section III.D. Besides, the VIO subsystem includes the LiDAR-depth enhancement. R3LIVE and FAST-LIVO2 are SOTA tightly-coupled LVIO, which are the baseline methods for the proposed PA-LVIO. The accuracy evaluation is conducted on the desktop PC, and all systems are tested in real-time mode.

#### 1) i2Nav-Robot Dataset

The absolute translation errors (ATEs) of the employed methods on the *i2Nav-Robot* dataset are shown in Table II. The proposed PA-LVIO yields superior results than SOTA odometry methods in terms of the average ATE. Besides, PA-LVIO exhibits the minimum ATEs on half of the sequences with decimeter-level accuracy on 8 sequences. Meanwhile, PA-LVIO also demonstrate better results on the *building* and *street* sequences, which are more challenging with large-scale scenarios. In contrast, the SOTA methods show larger ATEs on these sequences with a meter-level odometry accuracy.

The VIO and LIO subsystems of PA-LVIO also exhibit satisfactory results, especially for the VIO subsystem. By tightly integrating the accurate translation-measurement capability of LiDAR and rotation-measurement capability of camera, PA-LVIO yields improved accuracy on nearly all sequences. We notice that the ATEs of PA-LVIO are much

TABLE II

ABSOLUTE TRANSLATION ERRORS (RMSE, METERS) ON *I2NAV-ROBOT* AND *MARS-LVIG* DATASETS

| Dataset | Sequence | FF-LINS | FAST-LIO2 | LE-VINS | R3LIVE | FAST-LIVO2 | Ours (VIO) | Ours (LIO) | Ours (w/o F2M) | Ours (Marg. F2M) | Ours (w/o calib.) | Ours (PA-LVIO) |
|---|---|---|---|---|---|---|---|---|---|---|---|---|
| *i2Nav-Robot* | *building00* | 2.32 | 0.68 | 1.17 | × | 1.10 | 0.68 | 0.47 | 0.70 | 2.39 | **0.27** | 0.34 |
| | *building01* | 1.16 | 0.38 | 1.01 | × | 0.19 | 0.61 | 0.16 | 0.27 | 1.21 | **0.12** | 0.17 |
| | *building02* | 6.73 | 1.46 | 2.35 | 4.49 | 1.65 | 1.18 | 0.61 | 0.67 | 2.51 | 0.53 | **0.47** |
| | *parking00* | 1.94 | 0.25 | 0.98 | 0.59 | 0.26 | 0.59 | 0.18 | 0.57 | 0.19 | 0.12 | **0.11** |
| | *parking01* | 1.06 | 0.44 | 0.69 | 0.39 | 0.17 | 0.50 | 0.19 | 0.53 | 0.49 | 0.30 | **0.14** |
| | *parking02* | 1.31 | 1.46 | 0.79 | 0.66 | 0.18 | 0.51 | 0.13 | 0.55 | 0.53 | 0.17 | **0.11** |
| | *playground00* | 1.71 | 0.31 | 1.34 | 0.62 | 0.53 | 0.78 | 1.36 | 0.81 | 0.74 | 1.00 | **0.30** |
| | *street00* | 3.32 | 1.86 | 1.18 | 1.61 | 1.12 | 0.59 | 1.11 | 0.76 | 1.92 | **0.31** | 0.47 |
| | *street01* | 4.20 | 1.28 | 1.65 | 2.36 | 2.45 | **0.61** | 1.83 | 0.79 | 1.69 | 1.10 | 0.91 |
| | *street02* | 3.78 | 1.49 | 2.16 | 1.40 | 2.92 | **0.33** | 2.41 | 0.75 | 2.25 | 0.83 | 1.15 |
| | Average | 2.75 | 0.96 | 1.33 | 1.51 | 1.06 | 0.67 | 1.14 | 0.66 | 1.39 | 0.59 | **0.54** |
| *MARS-LVIG* | *AMtown01* | 3.35 | 2.99 | 13.72 | 1.67 | 2.57 | 19.31 | 3.16 | 3.38 | **1.4** | 2.96 | 3.05 |
| | *AMtown02* | 4.67 | 3.35 | 11.32 | **2.24** | 2.82 | 11.12 | 3.15 | 3.78 | 2.42 | 2.70 | 2.67 |
| | *AMtown03* | 2.53 | 3.54 | 31.23 | 3.7 | 3.25 | 26.58 | × | 3.79 | **1.36** | 2.15 | 1.56 |
| | *AMvalley01* | 1.77 | 6.72 | 12.92 | 3.91 | 7.78 | 4.63 | 4.06 | **1.05** | 4.15 | 3.92 | 2.05 |
| | *AMvalley02* | 2.15 | 7.75 | 12.74 | 3.98 | 3.55 | 6.22 | 2.12 | **1.05** | 2.83 | 2.36 | 2.01 |
| | *AMvalley03* | 3.62 | 11.98 | 18.88 | 3.95 | **1.43** | 7.02 | 5.41 | 2.57 | 4.95 | 4.59 | 2.77 |
| | *HKairport01* | 0.50 | 0.42 | 5.41 | 3.26 | 0.75 | 2.22 | × | 0.54 | **0.18** | 0.41 | 0.42 |
| | *HKairport02* | 0.95 | 0.67 | 4.15 | 1.48 | 0.88 | 2.08 | × | 0.57 | **0.23** | 0.82 | 0.52 |
| | *HKairport03* | 1.39 | 0.83 | 8.53 | 3.65 | 1.12 | 2.42 | × | 0.89 | **0.41** | 0.57 | 0.38 |
| | Average | 2.33 | 4.25 | 13.21 | 3.09 | 2.68 | 9.32 | 3.58 | 1.96 | 1.99 | 2.29 | **1.71** |

× denotes the system totally failed. The bold and underlined results denote the best and second best, respectively. The results of without using F2M (w/o F2M) and marginalizing F2M (Marg. F2M) are discussed in Section IV.B.4, while the results of without online calibration (w/o calib.) are introduced in Section IV.B.5.

TABLE III

END-TO-END ERRORS (METERS) ON *R3LIVE* DATASET

| Sequence | FF-LINS | FAST-LIO2 | LE-VINS | R3LIVE | FAST-LIVO2 | Ours (VIO) | Ours (LIO) | Ours (w/o F2M) | Ours (Marg. F2M) | Ours (w/o calib.) | Ours (PA-LVIO) |
|---|---|---|---|---|---|---|---|---|---|---|---|
| *main-building* | 1.20 | 1.38 | 0.96 | 0.11 | 1.27 | 0.72 | 0.12 | 0.96 | **0.06** | 0.87 | 0.08 |
| *campus00* | 2.41 | 5.29 | 11.09 | 5.20 | 5.96 | 14.26 | 2.21 | 3.07 | 6.14 | 2.55 | **0.10** |
| *campus01* | 2.51 | 2.19 | 2.78 | 19.57 | 5.68 | 5.27 | 1.35 | 3.03 | 4.22 | 9.29 | **0.66** |
| *campus02* | 4.03 | 0.08 | 2.38 | 0.08 | **0.01** | 2.55 | 0.13 | 2.69 | 0.55 | 1.91 | 0.18 |
| *park0* | 0.47 | 0.06 | 1.26 | 0.08 | **0.04** | 1.10 | 0.16 | 0.15 | 0.20 | 0.13 | 0.13 |
| *park1* | 1.45 | 0.56 | 0.89 | 0.60 | 0.54 | 0.75 | 0.52 | 0.65 | 0.63 | **0.50** | 0.58 |
| Average | 2.01 | 1.59 | 3.23 | 4.27 | 2.25 | 4.11 | 0.75 | 1.76 | 3.54 | 2.54 | **0.29** |

larger than the VIO system on the *street01* and *street02* sequences, mainly because of the complex multi-sensor fusion system. The state estimator is affected by many factors, such as sensor synchronization and dynamic objects, and thus the results are reasonable. Nevertheless, PA-LVIO shows the best result in terms of the average ATEs.

Fig. 5 depicts the attitude errors of the LVIO methods on the *street00* sequence. LE-VINS is a consistent system without involving F2M constraints, and thus roll and pitch angles are observable and gravity-aligned. Hence, the roll and pitch errors of LE-VINS are very small, while the yaw errors diverge gradually due to the unobservable characteristic. By incorporating the proposed marginalization-free F2M constraint, PA-LVIO shows similar roll and pitch errors to LE-VINS. Meanwhile, the yaw errors do not diverge, as the F2M constraint can eliminate drifts. The results illustrate that the negative impacts of the F2M constraint on PA-LVIO are limited, as mentioned in Section III.D.

*2) MARS-LVIG Dataset*

As only the low-rate real-time kinematic (RTK) results are originally provided as the ground truth positions in the *MARS-LVIG* dataset, we derive the high-rate ground-truth pose[1] using a post-processing RTK/INS software. According to the results in Table II, the *MARS-LVIG* dataset is much more challenging, as the UAV flies very fast with speeds ranging from 3 m/s to 12 m/s. Besides, the range of the cruising altitude is between 80 m (*AMtown* and *HKairport* sequences) and 130 m (*AMvalley* sequences), and thus the LiDAR and visual features are very far relative to the UAV.

As a result, the ATEs on the *MARS-LVIG* dataset are larger than those on the *i2Nav-Robot* dataset, as shown in Table II. Compared with the SOTA methods, PA-LVIO exhibits significantly improved accuracy, especially on larger-scale *AMtown* and *HKairport* sequences. LE-VINS shows the worst results on nearly all sequences, mainly because of the high cruising altitude, which is challenging for VIO methods.

The VIO and LIO subsystems exhibit worse results than PA-LVIO. The VIO subsystem shows similar results as LE-VINS due to the high cruising altitude. Meanwhile, the LIO subsystem fails on four sequences, because the UAV has very large rotation motions, which are challenging for the data association of LiDAR. Hence, PA-LVIO can achieve notably improved accuracy by tightly fusing the LiDAR and visual, as the LiDAR can provide accurate translation constraints, while the visual can provide accurate rotation constraints.

Fig. 6 depicts the trajectories on the *AMtown03* and *AMvalley03* sequences. According to the ATE results in Table II, PA-LVIO yields improved accuracy to FAST-LIVO2 on

[1]https://github.com/thlsealight/MARS-LVIG-GT

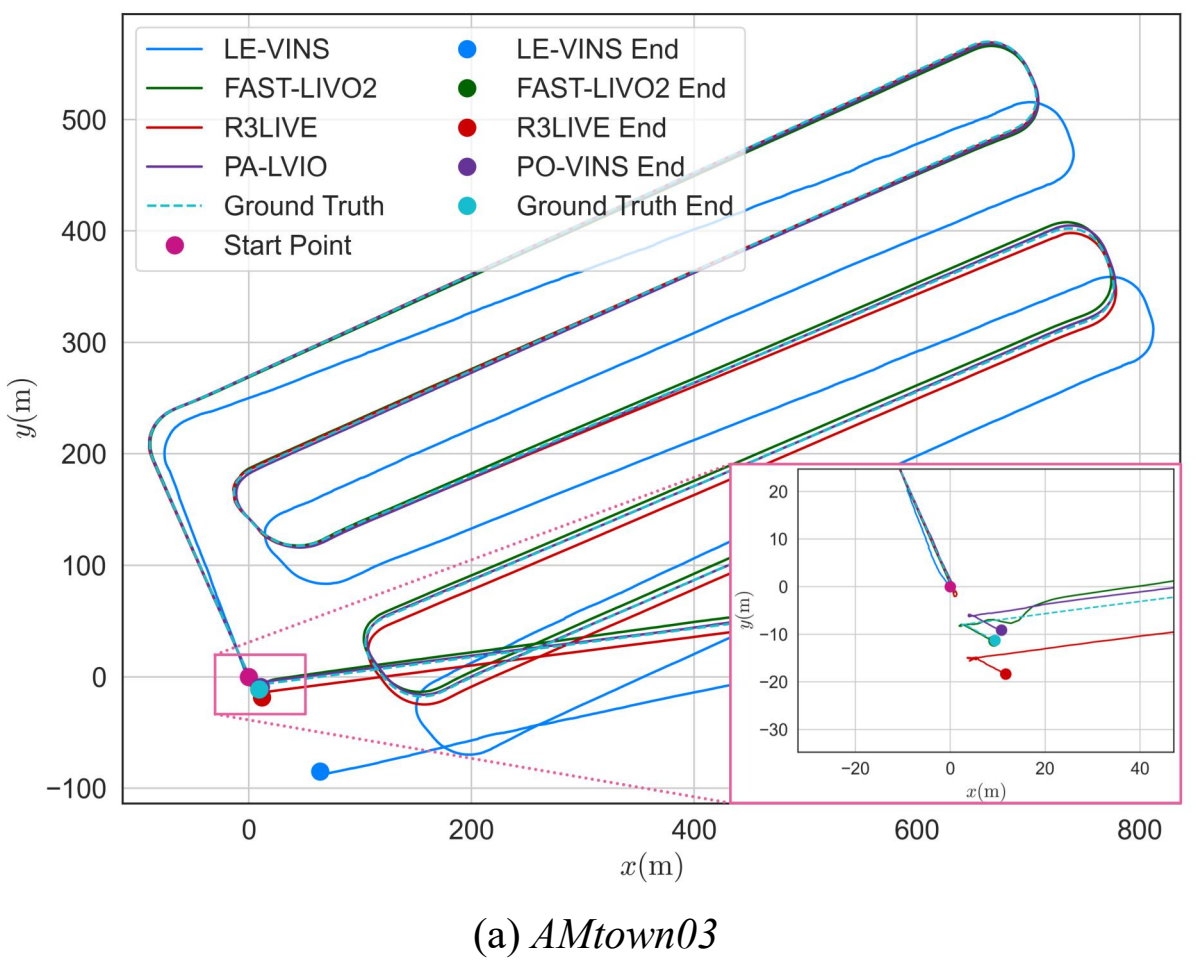

(a) *AMtown03*

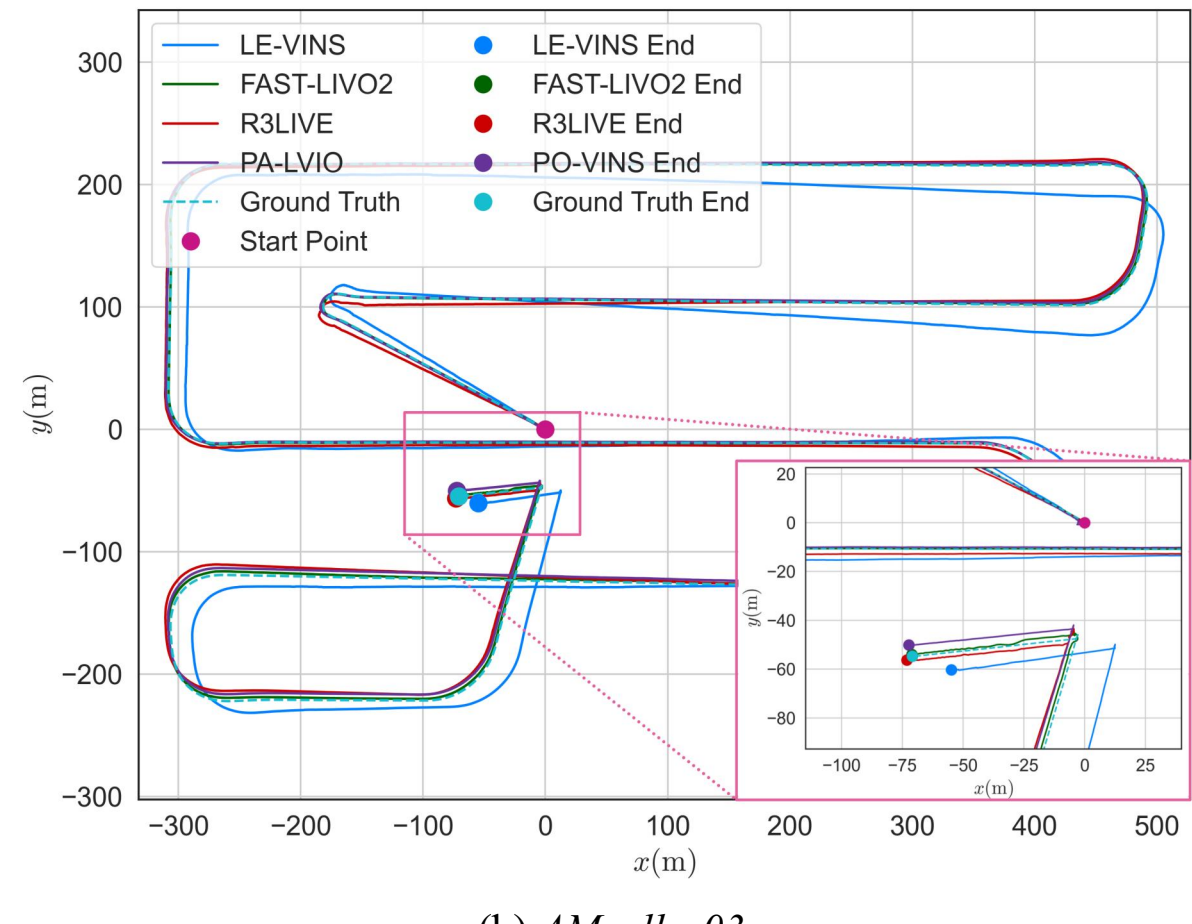

(b) *AMvalley03*

Fig. 6. The trajectory results on the *MARS-LVIG* dataset. The pink regions denote the trajectories at the end time, where there are trajectory jumps for FAST-LIVO2.

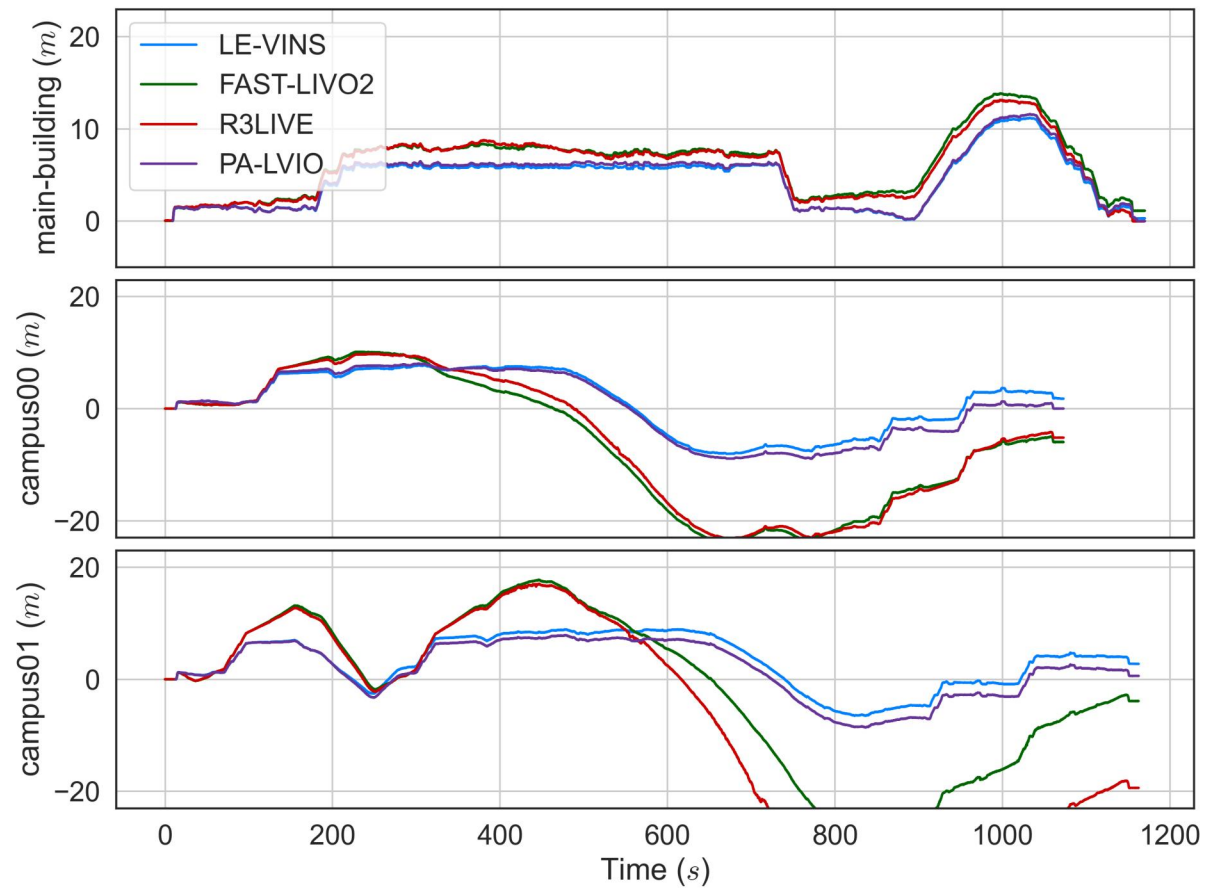

Fig. 7. Comparison of the height changes (along the $z$ axis) on the *R3LIVE* dataset.

*AMtown03* while degraded accuracy on *AMvalley03*. However, we notice that there are several jumps for FAST-LIVO2 at the end of the two sequences, as exhibited in Fig. 6. This is because FAST-LIVO2 matches with its previously built global map. In contrast, PA-LVIO shows a smoother trajectory without any jump, benefiting from the accuracy relative pose constraints from the F2F PA measurements. Besides, PA-LVIO demonstrates much better accuracy than R3LIVE and FAST-LVIO2 regarding the average ATEs.

*3) R3LIVE Dataset*

The end-to-end errors on the *R3LIVE* dataset are shown in Table III, and they are calculated by subtracting the start positions from the end positions. PA-LVIO also exhibits the best accuracy in terms of the average errors, and shows submeter-level end-to-end errors on all sequences. On the contrary, SOTA methods exhibit meter-level errors on some sequences, especially on the large-scale *campus00* and *campus01* with durations longer than 1,000 seconds. FAST-LIO2, R3LIVE, and FAST-LIVO2 achieve a centimeter-level accuracy on the *campus02* and *hku-park0*, which are two small-scale sequences with durations shorter than 500 seconds, while PA-LVIO exhibits a decimeter-level accuracy. The reason is that the proposed F2M pose constraints have not taken notable effect due to the short durations, as mentioned in III.D. Nevertheless, the differences of the end-to-end errors between PA-LVIO and SOTA methods are very small, and thus the results are acceptable.

Fig. 7 depicts the height changes on the *main-building*, *campus00*, and *campus01* sequences, in which there are significant height changes. The height of PA-LVIO nearly returns to zero at the end of the three sequences, while SOTA methods exhibit deviated height. Meanwhile, PA-LVIO shows smaller height changes between 200 s and 700 s on the *main-building*, because this segment of the dataset was collected in an indoor building with a flat floor. In contrast, R3LIVE and FAST-LIVO2 exhibit larger height changes even on the flat floor. The results indicate that PA-LVIO are gravity-aligned, and the negative impacts of the proposed F2M constraint are limited.

*4) Impact of the Marginalization-Free F2M Factor*

A LiDAR F2M pose factor is employed in PA-LVIO to eliminate odometry drifts, while it is not marginalized to reduce the negative impacts on the state estimator. As shown in Table II and Table III, PA-LVIO show improved accuracy than the method without using the F2M factor (expressed as w/o F2M) regarding the statistical results. More specifically, PA-LVIO yields notable improvement on the *building* and *parking* sequences of the *i2Nav-Robot* dataset, the *AMtown* sequences of the *MARS-LVIG* dataset, and the *campus* sequences of the *R3LIVE* dataset. Nevertheless, PA-LVIO exhibits degraded accuracy on some sequences, as the method without the F2M factor has already achieved superior accuracy. Meanwhile, the results also demonstrate that the negative impacts cannot be fully ignored.

Compared with the method marginalizing the F2M factor (expressed as Marg. F2M), PA-LVIO yield improved accuracy on all the three datasets in terms of the average errors, as shown in II and Table III. The results indicate that the proposed marginalization-free method can significantly reduce the negative influences of the F2M factor, especially on the *i2Nav-Robot* and *R3LIVE datasets*. We notice that PA-LVIO shows degraded accuracy on the *AMtown* sequences of the *MARS-LVIG* dataset. The reason is that the method without the F2M factor exhibits worse results, while the accuracy can improve notably by integrating the F2M factor. However, the effectiveness of the marginalization-free F2M factor in PA-LVIO is lower, though PA-LVIO also exhibits improved accuracy to the method without the F2M factor. All in all, the proposed marginalization-free F2M factor is an effective and applicable solution to eliminate odometry drifts.

TABLE IV
COMPARISON OF THE RUNNING TIME FOR DIFFERENT F2M METHODS ON THE *I2NAV-ROBOT* DATASET

| Sequence | Tightly coupled | Loosely coupled (PA-LVIO) | |
|---|---|---|---|
| | FGO (ms) | FGO (ms) | F2M Optimization (ms) |
| *building00* | 33.68 | 25.67 | 3.12 |
| *building01* | 31.01 | 23.53 | 3.03 |
| *building02* | 35.02 | 28.18 | 3.20 |
| *parking00* | 33.09 | 25.31 | 3.20 |
| *parking01* | 27.26 | 20.63 | 3.03 |
| *parking02* | 29.19 | 23.14 | 3.01 |
| *playground00* | 37.45 | 33.79 | 3.23 |
| *street00* | 37.36 | 30.69 | 3.29 |
| *street01* | 38.18 | 31.06 | 3.45 |
| *street02* | 37.23 | 27.64 | 3.34 |
| Average | 33.95 | 26.96 | 3.19 |

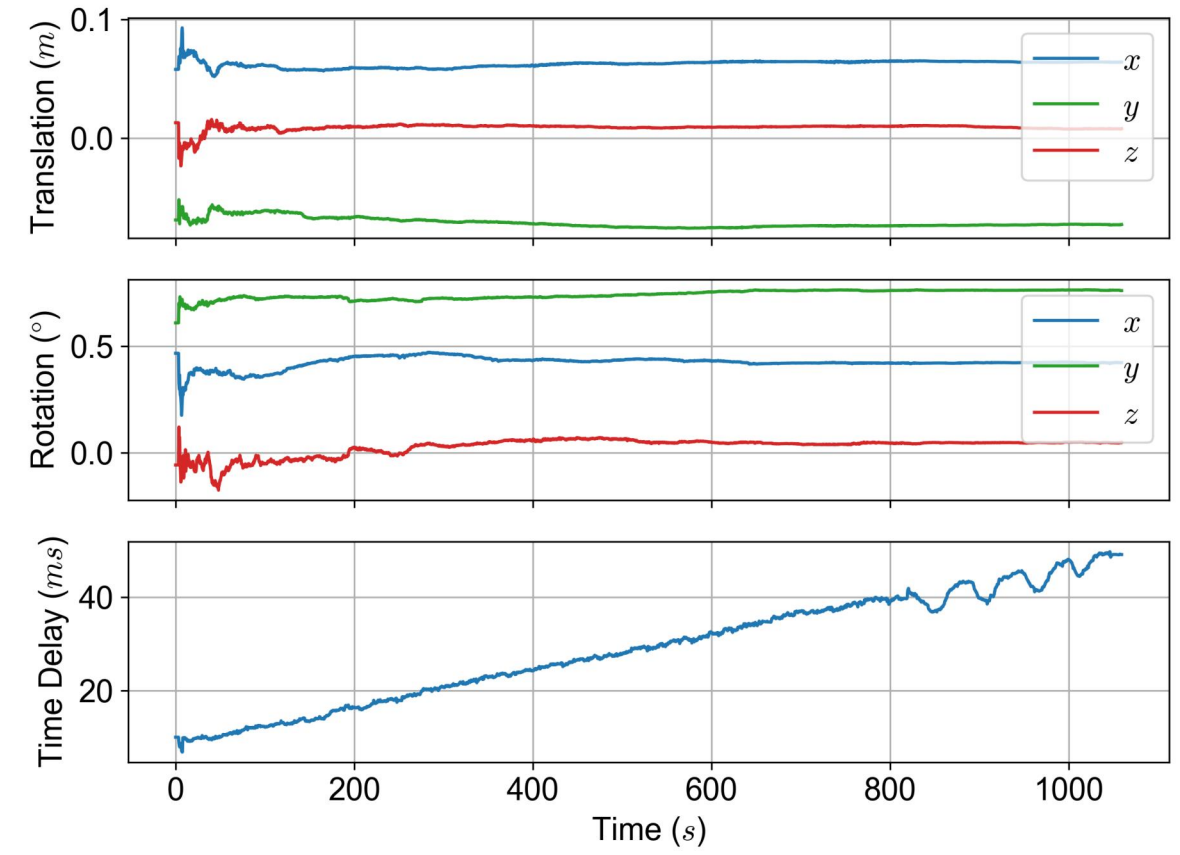

Fig. 8. Estimated camera-IMU spatial-temporal parameters on the *campus00* of the *R3LIVE* dataset. Here, only the tiny rotation parameters are reserved for better visualization.

We employ a loosely-coupled rather than a tightly-coupled F2M factor to improve the efficiency of the state estimation. The method using the tightly-coupled F2M factor exhibits nearly the same results as the PA-LVIO with centimeter-level differences, hence its results are not presented. According to the efficiency results in Table IV, the average FGO time of PA-LVIO is reduced by 20.6%, compared to the tightly-coupled method. Besides, the average time of the F2M pose optimization defined in (11) is only about 3 ms. Hence, the employed F2M pose factor is efficient while without affecting the odometry accuracy.

#### 5) *Impact of Online Calibration*

An IMU-centric online spatial-temporal calibration and compensation are employed to improve the robustness and accuracy of PA-LVIO. Fig. 8 exhibits the estimated camera-IMU spatial-temporal parameters on the *campus00* of the *R3LIVE* dataset. PA-LVIO not only can estimate the translation and rotation extrinsic parameters accurately, but also can estimate the time-varying camera-IMU temporal parameter. According to the results in Fig. 8, we can easily determine that the dataset was collected in indoor scenarios between 800 s and 1,000 s, due to the notably and frequently changed exposure time of the camera.

TABLE V
ESTIMATED ROTATION PARAMETERS BETWEEN LIDAR AND CAMERA ON THE *I2NAV-ROBOT* DATASET

| Sequence | *x* (°) | *y* (°) | *z* (°) |
|---|---|---|---|
| *building00* | -90.611 | 0.266 | 269.836 |
| *building01* | -90.596 | 0.255 | 269.854 |
| *building02* | -90.586 | 0.251 | 269.860 |
| *parking00* | -90.586 | 0.243 | 269.879 |
| *parking01* | -90.677 | 0.245 | 269.843 |
| *parking02* | -90.694 | 0.249 | 269.852 |
| *playground00* | -90.568 | 0.231 | 269.866 |
| *street00* | -90.611 | 0.257 | 269.872 |
| *street01* | -90.600 | 0.249 | 269.869 |
| *street02* | -90.583 | 0.248 | 269.897 |
| Mean[1] | -90.610 ±0.039 | 0.250 ±0.009 | 269.860 ±0.017 |

[1]The mean value and the standard deviation.

According to the results in Table II and Table III, PA-LVIO yields improved accuracy than the method without online calibration (expressed as w/o calib.), especially on the *MARS-LVIG* and *R3LIVE* datasets. As accurate extrinsic parameters are provided in the *i2Nav-Robot* dataset, the accuracy improvement for PA-LVIO is limited. Besides, PA-LVIO yields significantly improved accuracy on the *campus00* and *campus01* of the *R3LIVE* dataset, as shown in Table III. The reason is that the camera-IMU temporal parameters are changed severely due to the inaccurate sensor synchronization, as shown in Fig. 8.

### C. *Evaluation of Mapping*

#### 1) *Evaluation of LiDAR-Camera Alignment*

Accurate LiDAR-camera alignment should be achieved to derive RGB-rendered point-cloud maps. As the rotation parameters between the LiDAR and camera play a more important role for accurate mapping, we evaluate the rotation parameters on the *i2Nav-Robot* dataset. According to the results in Table V, the STDs of the estimated rotation parameters for three angles are all less than 0.04°, demonstrating superior accuracy. The pixel size of the employed camera in the *i2Nav-Robot* dataset is 5.86 um, and the focal length is 6 mm, according to their provided datasheets. The angle corresponding to one pixel can be approximately calculated as $\alpha = \sin^{-1} 5.86\mathrm{um}/6\mathrm{mm} = 0.0573°$. As a result, the estimated LiDAR-camera rotation parameters are with sub-pixel-level accuracy. We can still achieve pixel-wise LiDAR-camera alignment, when other errors, such as LiDAR-camera translation parameters, are considered.

#### 2) *Evaluation of Mapping Quality*

We qualitatively evaluate the mapping results of PA-LVIO on different datasets. Three new sequences collected by the *HandNav* device shown in Fig. 4 are adopted for a detailed comparison, including the *whu-building*, *whu-gateway*, and *whu-library* sequences. The SOTA LiDAR-visual-inertial odometry and mapping methods R3LIVE and FAST-LVIO2 are employed as the baseline systems.

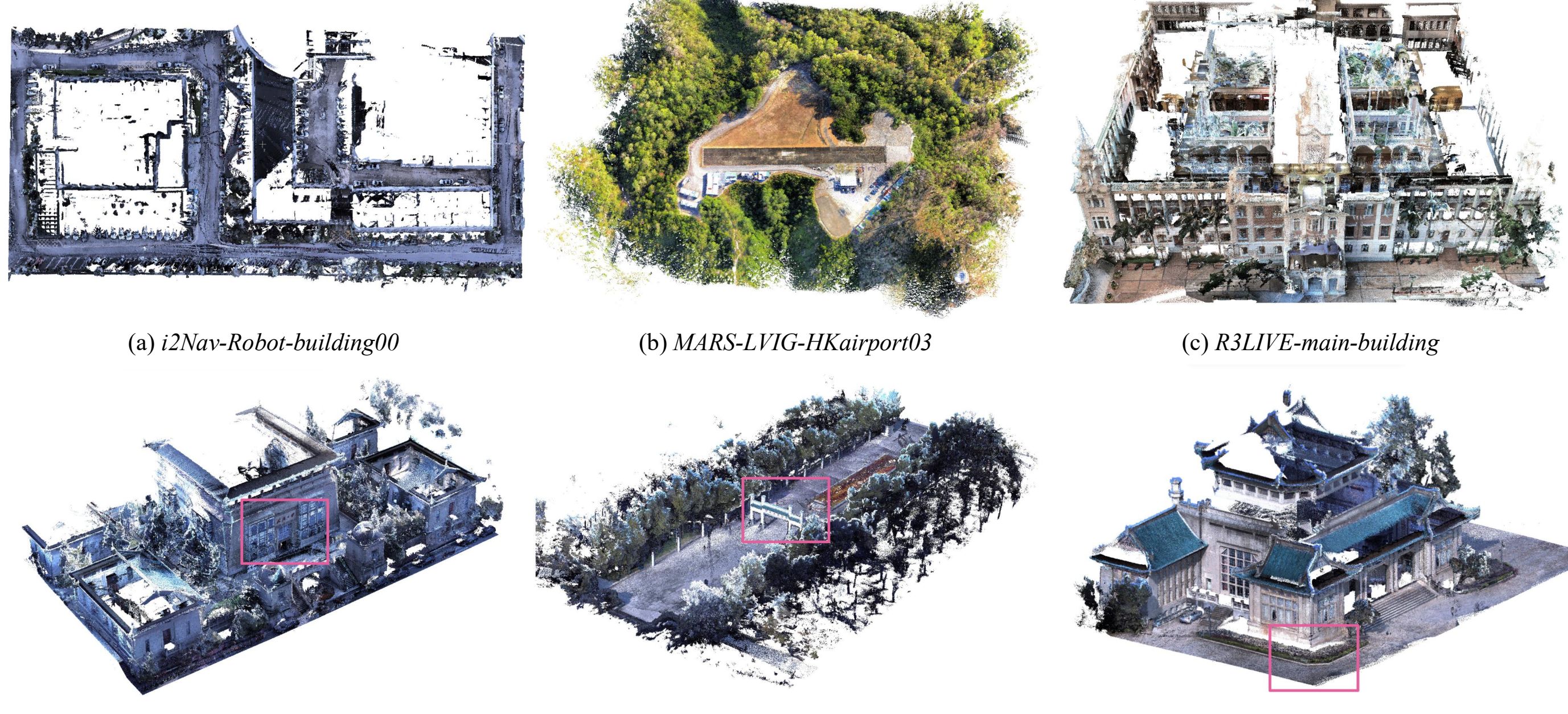

(a) *i2Nav-Robot-building00* (b) *MARS-LVIG-HKairport03* (c) *R3LIVE-main-building*

(d) *whu-building* (e) *whu-gateway* (f) *whu-library*

Fig. 9. Mapping results from PA-LVIO. (a) and (b) are shown in the top view. These maps have been cut for better visualization. The pink rectangles denote the local regions in Fig. 10.

Fig. 9 depicts the mapping results of the proposed PA-LVIO on different datasets. We can easily distinguish the road signs and vehicles in Fig. 9.a, though there are many noisy points caused by the dynamic objects on the roads. The runway, trees, and buildings in Fig. 9.b are also very clear, which demonstrates the superior mapping performance of PA-LVIO. Meanwhile, the main building of HKUST is well reconstructed by PA-LVIO, and the point-cloud colors are correctly rendered. The built point-cloud maps on the three sequences of the *HandNav* dataset can clearly restore the textures of the traditional buildings in Wuhan University, as exhibited in Fig. 9.d~Fig. 9.f. The results indicate that the state estimator and the spatial-temporal alignment in PA-LVIO are accurate to build high-quality and RGB-rendered point-cloud maps.

The detailed mapping results of the *whu-building*, *whu-gateway*, and *whu-library* sequences are shown in Fig. 10. Compared to R3LIVE and FAST-LIVO2, PA-LVIO yields the best mapping results in the three sequences. Specifically, R3LIVE and FAST-LIVO2 exhibit distorted and blurred textures on the *whu-building*, as the structures of the building are not straight. On the *whu-gateway*, the maps built by SOTA methods have notable drifts regarding the archway, mainly because of the unstructured scenarios, as shown in Fig. 9.e. Besides, it is difficult to distinguish the tiles on the *whu-library* for SOTA methods. In contrast, PA-LVIO exhibits clearer point-cloud maps and less drifts, benefiting from the accurate odometry which integrates both F2F and F2M constraints. Meanwhile, PA-LVIO shows more distinguished textures, such as the clear characters and tiles, which thanks to the employed online spatial-temporal calibration and compensation. Consequently, we can achieve pixel-wise LiDAR-camera alignment for high-quality mapping.

TABLE VI
COMPARISON OF PROCESSING TIME ON DIFFERENT PLATFORMS

| Sequence | LiDAR (ms) | | Visual (ms) | | FGO (ms) | |
|---|---|---|---|---|---|---|
| | PC | ARM | PC | ARM | PC | ARM |
| *whu-building* | 2.8 | 18.2 | 7.28 | 29.34 | 33.96 | 77.12 |
| *whu-gateway* | 3.6 | 20.6 | 7.01 | 28.45 | 30.56 | 66.29 |
| *whu-library* | 3.4 | 22.4 | 6.92 | 28.95 | 52.66 | 112.64 |
| Average | 3.27 | 20.40 | 7.07 | 28.91 | 39.06 | 85.35 |

The LiDAR and visual processes mainly include the preprocessing and data associations.

TABLE VII
TOTAL RUNNING TIME AND THE EQUIVALENT FPS

| Sequence | Sequence length (s) | Total running time (s) | | Equivalent FPS (Hz) | |
|---|---|---|---|---|---|
| | | PC | ARM | PC | ARM |
| *whu-building* | 622 | 169 | 427 | 36.8 | 14.6 |
| *whu-gateway* | 313 | 89 | 240 | 35.2 | 13.0 |
| *whu-library* | 426 | 154 | 378 | 27.7 | 11.3 |
| Average | × | × | × | 33.2 | 13.0 |

The equivalent FPS is calculated by dividing the sequence length by the running time and multiplying by the frame rate.

## D. *Evaluation of Efficiency*

We evaluate the efficiency of PA-LVIO on the desktop PC (AMD Ryzen 9-9950X CPU and NVIDIA RTX 5080 GPU) and the onboard ARM computer (NVIDIA Orin NX with 6-core CPU and 8-GB RAM) on the HandNav dataset. The original image with a resolution of 1800*1200 is resized to 900*600 on the onboard computer for better real-time performance, due to the limited computational resources.

The processing time of PA-LVIO on different platforms is shown in Table VI. As the desktop PC has more CPU cores, PA-LVIO runs much faster on the desktop PC than the onboard computer by using multi-threading technology. Note that only the LiDAR preprocessing, visual preprocessing, and visual feature tracking should be conducted frame by frame. Hence, PA-LVIO can run in real-time on both the two platforms, though the FGO process consumes lots of computational resources to some extent. Meanwhile, the IMU can provide high-rate pose by INS mechanization for applications like robot navigation and planning.

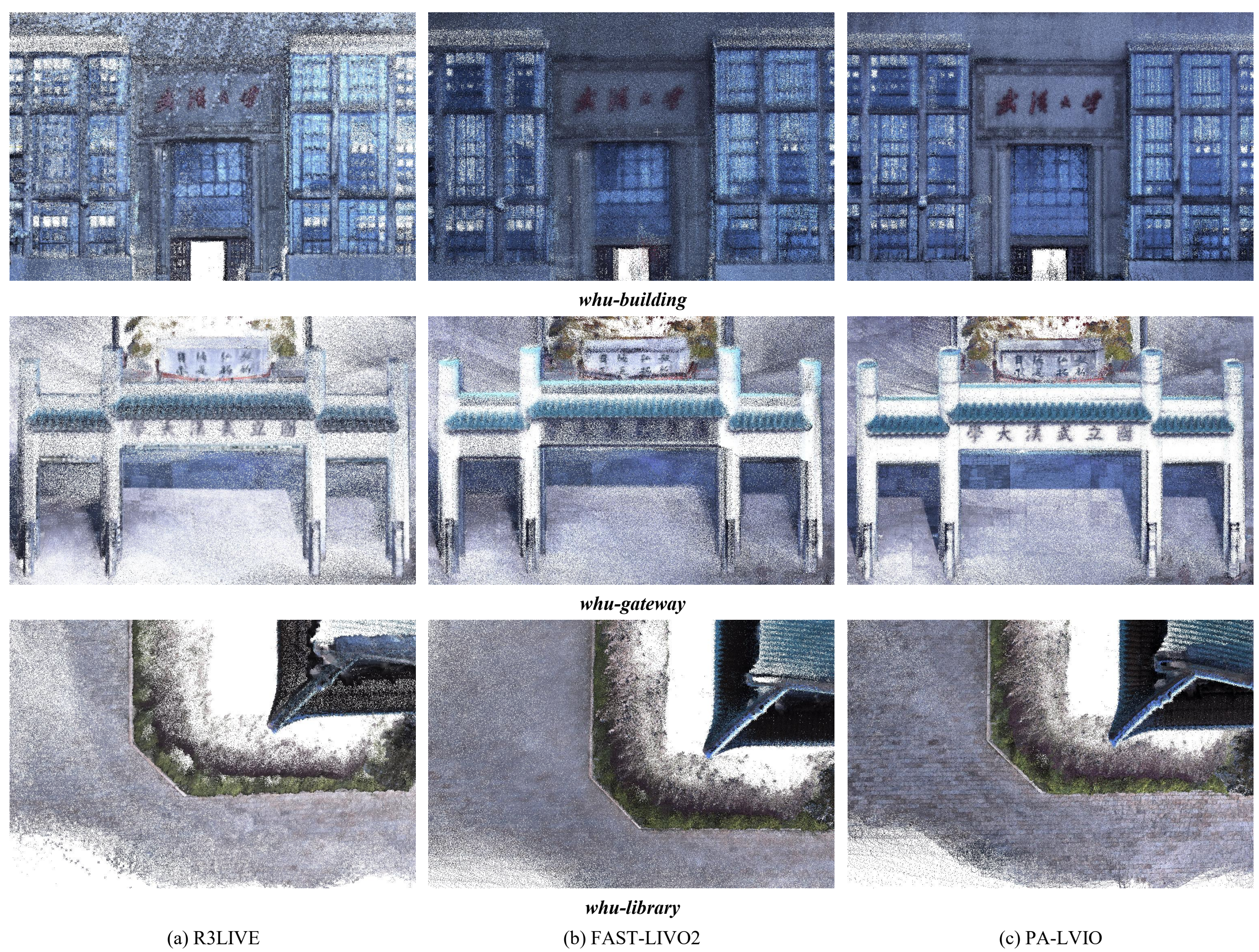

Fig. 10. Comparison of detailed mapping results from different methods. Each column denotes the results from each method.

The statistical results of the total running time and the equivalent frame per second (FPS) are exhibited in Table VII. Here, the equivalent FPS is calculated by dividing the sequence length by the running time and multiplying by the frame rate. The total running time of PA-LVIO is far less than the sequence lengths, especially on the desktop PC. According to the average FPS results, PA-LVIO can run at more than 33 Hz on the desktop PC, while 13 Hz on the onboard desktop PC. The results demonstrate that PA-LVIO is efficient and can run in real time on the onboard ARM computer with fewer computational resources and memory, even without any tuning.

## V. Conclusions

In this study, we propose PA-LVIO, a real-time LiDAR-visual-inertial odometry and mapping system, with a novel pose-only bundle adjustment (PA) framework. The LiDAR and visual frame-to-frame (F2F) PA measurements together with the IMU-preintegration measurements are tightly integrated in a unified state estimator for accurate and consistent dead reckoning. Besides, we employ the marginalization-free LiDAR frame-to-map (F2M) measurements to eliminate the odometry drifts, while mitigating the negative effects on the state estimator. The spatial-temporal parameters are delicately modeled and incorporated into the estimator for online calibration and compensation. Hence, we can build high-definition and RGB-rendered point-cloud maps. Sufficient experiments on both the public and private datasets indicate that the proposed PA-LVIO yields superior or comparable odometry accuracy and mapping quality to SOTA methods. Meanwhile, PA-LVIO demonstrates satisfied real-time performance, even on the onboard ARM computer.

Nevertheless, we notice that the marginalization-free F2M

measurement may still reduce the odometry accuracy in certain circumstances, which means its shortcomings cannot be completely ignored. Besides, the FGO costs too much computational resources, though the employed PA measurements are very efficient. Hence, we will try to further improve the F2M measurement model to eliminate its negative impacts on the estimator. In the future, we will also integrate the F2F PA measurement models into an MSCKF-based estimator for more efficient LVIO.